\documentclass[twoside,11pt]{article}

\usepackage{blindtext}
\usepackage{dsfont}
\usepackage{mathalfa, mathrsfs}
\usepackage{amsfonts,mathtools}
\usepackage[noend]{algorithmic}
\usepackage[ruled,vlined,linesnumbered]{algorithm2e}
\usepackage{tikz, subcaption}
\usepackage{xspace}
\usetikzlibrary{positioning}
\usepackage{adjustbox}
\usepackage[inline]{enumitem}
\usepackage{booktabs}
\usepackage{siunitx}

\usetikzlibrary{shapes, arrows, arrows.meta, positioning}
\usetikzlibrary{shapes.geometric, calc}

\definecolor{bg}{gray}{0.95}

\DeclareMathOperator*{\argmin}{arg\,min}
\DeclareMathOperator*{\argmax}{arg\,max}

\newcommand{\independent}{\perp\!\!\!\perp}
\newcommand{\notindependent}{\not\!\perp\!\!\!\perp}
\newcommand{\sep}[4]{\left(#1 \independent #2 \vert #3\right)_{#4}}
\newcommand{\nsep}[4]{\left(#1 \notindependent #2 \vert #3\right)_{#4}}

\newcommand{\Pa}[2]{\textit{Pa}_{#2}(#1)}
\newcommand{\Ch}[2]{\textit{Ch}_{#2}(#1)}
\newcommand{\Ne}[2]{\textit{Ne}_{#2}(#1)}
\newcommand{\Anc}[2]{\textit{Anc}_{#2}(#1)}

\newcommand{\Dis}[2]{\textit{Dis}_{#2}(#1)}
\newcommand{\PaP}[2]{\textit{Pa}_{#2}^+(#1)}
\newcommand{\CP}[2]{\Lambda_{#2}(#1)}
\newcommand{\VS}[2]{\text{VS}_{#2}(#1)}
\newcommand{\Mb}[2]{\text{Mb}_{#2}(#1)}
\newcommand{\MB}[1]{\text{Mb}_{#1}}

\newcommand{\D}[0]{\text{Data}}

\newcommand{\Vb}[0]{\mathbf{V}}

\newcommand{\Wb}[0]{\mathbf{W}}
\newcommand{\Ub}[0]{\mathbf{U}}
\newcommand{\Eb}[0]{\mathbf{E}}
\newcommand{\Xb}[0]{\mathbf{X}}

\newcommand{\Zb}[0]{\mathbf{Z}}

\newcommand{\Sb}[0]{\mathbf{S}}

\newcommand{\PV}[0]{P_{\Vb}}

\newcommand{\G}[0]{\mathcal{G}}

\newcommand{\F}{\textsc{SkipCheck\_vec}\xspace}
\newcommand{\FF}{\textsc{SkipCheck\_mat}\xspace}
\newcommand{\FI}{\textsc{SkipCheck\_cond1}\xspace}
\newcommand{\FII}{\textsc{SkipCheck\_cond2}\xspace}

\usepackage{multirow}
\usepackage{array}
\newcolumntype{M}[1]{>{\centering\arraybackslash}m{#1}}
\newcolumntype{N}{@{}m{0pt}@{}}

\DeclareMathAlphabet\mathbfcal{OMS}{cmsy}{b}{n}

\newcommand{\GV}[0]{\G_{\Vb}}
\newcommand{\Gpi}{\G^{\pi}}

\newcommand{\true}[0]{\textsc{True}\xspace}
\newcommand{\false}[0]{\textsc{False}\xspace}
\newcommand{\sepset}[0]{\text{SepSet}\xspace}

\newcommand{\R}[1]{$\mathfrak{R}_{#1}$\label{R#1}}
\newcommand{\refR}[1]{\hyperref[R#1]{$\mathfrak{R}_{#1}$}}

\usepackage{lastpage}

\usepackage{jmlr2e}

\ShortHeadings{Recursive Causal Discovery}{Mokhtarian, Elahi, Akbari, and Kiyavash}
\firstpageno{1}

\begin{document}

\title{Recursive Causal Discovery}

\author{\name Ehsan Mokhtarian \email ehsan.mokhtarian@epfl.ch \\
       \addr School of Computer and Communication Sciences\\
        EPFL, Lausanne, Switzerland\\
       \AND
       \name Sepehr Elahi \email sepehr.elahi@epfl.ch \\
       \addr School of Computer and Communication Sciences\\
        EPFL, Lausanne, Switzerland\\
       \AND
       \name Sina Akbari \email sina.akbari@epfl.ch \\
       \addr School of Computer and Communication Sciences\\
        EPFL, Lausanne, Switzerland\\
       \AND
       \name Negar Kiyavash \email negar.kiyavash@epfl.ch \\
       \addr College of Management of Technology\\
        EPFL, Lausanne, Switzerland\\
       }

\editor{ }

\maketitle

\begin{abstract}
    Causal discovery, i.e., learning the causal graph from data, is often the first step toward the identification and estimation of causal effects, a key requirement in numerous scientific domains. Causal discovery is hampered by two main challenges: limited data results in errors in statistical testing and the computational complexity of the learning task is daunting. This paper builds upon and extends four of our prior publications \citep{mokhtarian2021recursive, akbari2021recursive, mokhtarian2022learning, mokhtarian2023novel}. These works introduced the concept of \emph{removable} variables, which are the only variables that can be removed recursively for the purpose of causal discovery. Presence and identification of removable variables allow recursive approaches for causal discovery, a promising solution that helps to address the aforementioned challenges by reducing the problem size successively. This reduction not only minimizes conditioning sets in each conditional independence (CI) test, leading to fewer errors but also significantly decreases the number of required CI tests. The worst-case performances of these methods nearly match the lower bound. In this paper, we present a unified framework for the proposed algorithms, refined with additional details and enhancements for a coherent presentation. A comprehensive literature review is also included, comparing the computational complexity of our methods with existing approaches, showcasing their state-of-the-art efficiency. Another contribution of this paper is the release of \texttt{RCD}, a Python package that efficiently implements these algorithms. This package is designed for practitioners and researchers interested in applying these methods in practical scenarios. The package is available at \href{https://github.com/ban-epfl/rcd}{\url{github.com/ban-epfl/rcd}}, with comprehensive documentation provided at \href{https://rcdpackage.com}{\url{rcdpackage.com}}.
\end{abstract}



\section*{Note to Readers}
    This paper builds upon and extends the methodologies from four of our previous publications in causal discovery. These publications are outlined as follows:
    \begin{itemize}
        \item[\R{1}]\cite{mokhtarian2021recursive}: A recursive Markov boundary-based approach to causal structure learning; published in KDD-CD 2021; introducing \textbf{MARVEL}.
        \item[\R{2}] \cite{akbari2021recursive}: Recursive causal structure learning in the presence of latent variables and selection bias; published in NeurIPS 2021; introducing \textbf{L-MARVEL}.
        \item[\R{3}] \cite{mokhtarian2022learning}: Learning Bayesian networks with structural side information; published in AAAI 2022; introducing \textbf{RSL}.
        \item[\R{4}] \cite{mokhtarian2023novel}: Novel ordering-based approaches for causal structure learning in the presence of unobserved variables; published in AAAI 2023; introducing \textbf{ROL}.
    \end{itemize}
    For readers primarily interested in the practical implementation of our proposed methods, we recommend proceeding directly to Section \ref{sec: RCD}, where our Python package \texttt{RCD} is introduced.
    Additionally, to facilitate ease of reading, we have summarized the key notations in Table \ref{table of notations}.
    The main results discussed in this paper are also summarized in Table \ref{table: summary of results}.

\section{Introduction} \label{sec: intro}
    A fundamental task in various fields of science is discovering the causal relations among the variables of interest in a system.
    These relations, pivotal for tasks such as inference and policy design, are often encoded as a \emph{maximal ancestral graph} (MAG), representing the \emph{causal structure} of the system.
    In cases where no hidden variables are present, a MAG simplifies to a \emph{directed acyclic graph} (DAG).
    The problem of learning the causal structure of a system, known as the causal discovery problem, is notoriously challenging and is recognized as an NP-hard problem \citep{chickering2004large}.

    To address this problem, a slew of methodologies, broadly categorized into \emph{score-based} and \emph{constraint-based} methods, have been developed.
    Score-based methods learn the graph as a solution of an optimization problem that maximizes a predefined score.
    On the other hand, constraint-based methods leverage statistical tests to identify structures consistent with the \emph{conditional independence} (CI) relations observed in data.
    Additionally, \emph{hybrid} methods have emerged that combine both constraint- and score-based methodologies.
    This paper primarily focuses on constraint-based methods, which are most commonly used in the presence of hidden variables.
    In Section \ref{sec: related work}, we present a comprehensive review of existing causal discovery methods.

    Causal discovery is plagued by numerous challenges, most critically those pertaining to computational/time complexity and sample efficiency.
    In constraint-based methods, time complexity is primarily determined by the number of required CI tests.
    The two classical approaches, the PC algorithm, developed for DAG-learning, and its counterpart for MAG-learning, the FCI algorithm, are not scalable to large graphs \citep{spirtes2000causation}.
    Subsequent research has focused on reducing the computational burden and improving the statistical efficiency of these seminal works.
    On the other hand, methods such as the RFCI \citep{colombo2012learning}, specifically designed to gain computational speed, do this at the cost of possibly compromising completeness.
    Completeness refers to a method's asymptotic correctness, i.e., when sufficiently large numbers of samples are available so that the statistical tests used in the method are error-free.

    Recent advancements, such as our four publications (\refR{1}-\refR{4}) in recursive causal discovery, have made significant strides in addressing time and sample complexity while maintaining completeness guarantees.
    Our proposed framework strategically identifies a so-called \emph{removable} variable (Definition \ref{def: removable}), denoted by $X$, and learns its neighbors.
    After omitting this variable, the causal structure over the remaining variables is learned using only the samples from those variables.
    It is essential to carefully select $X$ to avoid any erroneous inclusions or exclusions in the causal graph, which we shall explain in detail with examples in Section \ref{sec: theory}.
    As these methods operate iteratively, they systematically reduce the problem size, leading to fewer CI tests and smaller conditioning sets; hence improving the statistical reliability of these tests.
    
    \begin{table}[t!]
        \centering
        \resizebox{\textwidth}{!}{
        \begin{tabular}{ccc|cc|c|c}
            \toprule
            &\multicolumn{2}{c|}{\textbf{Algorithm}} & \multicolumn{2}{c|}{\textbf{Assumptions}}  & \multirow{2}{*}{\textbf{Completeness}}& \multirow{2}{*}{\textbf{\#CI tests}}
            \\
            & Reference
            & Name
            & Causal sufficiency
            & Other
            & 
            & 
            \\
            \hline
            & \refR{1}
            & MARVEL
            & YES
            & -
            & YES
            & $\mathcal{O}(n^2 + n\Delta_{in}^2 2^{\Delta_{in}})$
            \\
            [0.5ex]
            \hline
            & \refR{2}
            & L-MARVEL
            & NO
            & -
            & YES
            & $\mathcal{O}(n^2 + n ({\Delta_{in}^+})^2 2^{\Delta_{in}^+})$
            \\
            [0.5ex]
            \hline
            & \multirow{2}{*}{\textbf{\refR{3}}}
            & RSL$_\omega$
            & YES
            & $\omega(\G) \leq m$
            & YES
            & $\mathcal{O}(n^2 + n\Delta_{in}^{m+1})$
            \\
            & 
            & RSL$_D$
            & YES
            & Diamond-free
            & YES
            & $\mathcal{O}(n^2 + n\Delta_{in}^3)$
            \\
            [0.5ex]
            \hline
            & \multirow{3}{*}{\textbf{\refR{4}}}
            & ROL$_{\text{HC}}$
            & NO
            & -
            & NO
            & $\mathcal{O}(\textsc{maxIter} \times n^3)$
            \\
            & 
            & ROL$_{\text{VI}}$
            & NO
            & -
            & YES
            & $\mathcal{O}(n^2 2^n)$
            \\
            &
            & ROL$_{\text{PG}}$
            & NO
            & -
            & NO
            & N/A
            \\
            [0.5ex]
            \hline
            & \multicolumn{2}{c|}{\textbf{Lower Bound}}
            & YES
            & -
            & YES
            & $\Omega(n^2 + n\Delta_{in} 2^{\Delta_{in}})$
            \\
            & \multicolumn{2}{c|}{\textbf{Lower Bound}}
            & NO
            & -
            & YES
            & $\mathcal{O}(n^2 + n {\Delta_{in}^+} 2^{\Delta_{in}^+})$
            \\
            \bottomrule
        \end{tabular}
        }
        \caption{Summary of the assumptions, guarantees, and complexities of the recursive causal discovery methods discussed in the paper. For a list of causal discovery algorithms in the literature, refer to Tables \ref{table: all algorithms} and \ref{table:related}.}
        \label{table: t1}
    \end{table}

    Table \ref{table: t1} provides a concise summary of the assumptions, guarantees, and complexity of the recursive causal discovery methods discussed in this paper\footnote{
    For the notations used in  Table \ref{table: t1}, please refer to Table \ref{table of notations}.}.
    In this table, causal sufficiency refers to when all variables in the system are observable.
    The last column shows the number of total CI tests performed as a common proxy to measure the complexity of constraint-based methods.
    As mentioned earlier, causal discovery from observational data is NP-hard \citep{chickering2004large}.
    In the last two rows, we present lower bounds for the complexity of constraint-based methods under causal sufficiency and in the absence of it, as established in Section \ref{sec: lower bound}.
    We shall discuss in more detail in Section \ref{sec: complexity} how this table demonstrates the state-of-the-art efficiency of our proposed methods under various assumptions.

    Our main contributions in this paper are as follows.
    \begin{itemize}
        \item We present a unified framework for the algorithms proposed in \refR{1}-\refR{4}, refined with additional details and enhancements for a coherent presentation.
        \item We launch the \texttt{RCD} Python package, an efficient implementation of our recursive algorithms.
        \item We conduct a comprehensive literature review and compare the computational efficiency of our methods with existing approaches.
    \end{itemize}

    The remainder of the paper is organized as follows.
    We provide preliminaries, including formal definitions and the goal of causal discovery from observational data in Section \ref{sec: preliminaries}.
    The theoretical foundations of our proposed algorithms are established in Section \ref{sec: theory}.
    In this section, we cover the generic recursive framework for causal discovery, describe the concept and characteristics of removable variables, and provide a comparison of our novel removable orders with traditional approaches.
    Section \ref{sec: RCD methods} introduces four recursive causal discovery methods: MARVEL, L-MARVEL, RSL, and ROL, and discusses their integration within the recursive framework.
    In Section \ref{sec: implementation details}, we delve into the implementation details of these methods and provide detailed pseudocode for each of the aforementioned algorithms.
    Section \ref{sec: complexity} discusses the complexity and completeness of various causal discovery methods, with a particular emphasis on our proposed recursive approaches, alongside lower bounds that provide theoretical limits for constraint-based methods.
    An extensive literature review is carried out in Section \ref{sec: related work}.
    Finally, Section \ref{sec: RCD} is dedicated to introducing our Python package \texttt{RCD}, which efficiently implements our proposed recursive causal discovery algorithms.

\begin{table}[p]
    \centering
    \begin{adjustbox}{width=0.98\textwidth, totalheight=0.98\textheight, keepaspectratio}
        \begin{tabular}{N M{2.3cm}|M{9.5cm}|M{2.2cm}}
            \toprule
            & Notation & Description & Definition\\
            \hline
            & $\Vb$, $\Ub$ & Sets of observed and unobserved variables& \\
            & $n$ & Number of observed variables, i.e., $|\Vb|$& \\
            & MG & Mixed graph & \\
            & MAG & Maximal ancestral graph & \\
            & DAG & Directed acyclic graph & \\
            & $\Pa{X}{\G}$ & Parents of vertex $X$ in MG $\G$& \\
            & $\Pa{\Xb}{\G}$ & $\bigcup_{X \in \Xb}\Pa{X}{\G}$& \\
            & $\Ch{X}{\G}$ & Children of vertex $X$ in MG $\G$& \\
            & $\Ne{X}{\G}$ & Neighbors of vertex $X$ in MG $\G$& \\
            & $\Ne{X}{\Wb}$& $\{Y \in \Wb\setminus \{X\} \vert \forall \Zb \subseteq \Wb \setminus \{X,Y\}:\: \nsep{X}{Y}{\Zb}{P_{\Wb}}\}$ & Definition \ref{def: Neighbors}\\
            & $\Anc{X}{\G}$ & Ancestors of vertex $X$ (including $X$) in MG $\G$& \\
            & $\Anc{\Xb}{\G}$ & $\bigcup_{X \in \Xb}\Anc{X}{\G}$& \\
            & $\CP{X}{\G}$ & Co-parents of vertex $X$ in DAG $\G$& Definition \ref{def: CP}\\
            & $\Mb{X}{\Wb}$ & Markov boundary of vertex $X$ with respect to $\Wb$& Definition \ref{def: Mb 1}\\
            & $\MB{\Wb}$ & $\{\Mb{X}{\Wb} \vert X \in \Wb \}$ & Definition \ref{def: Mb 1}\\
            & $\Mb{X}{\G}$ & Markov boundary of vertex $X$ in MG $\G$& Definition \ref{def: Mb 2}\\
            & $\VS{X}{\G}$ & V-structures in DAG $\G$ in which vertex $X$ is a parent & Definition \ref{def: VS}\\
            & $\Dis{X}{\G}$ & District set of vertex $X$ in MG $\G$ & Definition \ref{def: pap}\\
            & $\PaP{X}{\G}$ & $\Pa{X}{\G} \cup \Dis{X}{\G} \cup \Pa{\Dis{X}{\G}}{\G}$ & Definition \ref{def: pap}\\
            & $\Pi(\Wb)$ & Orders over $\Wb$ & Definition \ref{def: order}\\
            & $\Pi^c(\G)$ & c-orders of DAG $\G$ & Definition \ref{def: c-order}\\
            & $\Pi^r(\G)$ & r-orders of MAG $\G$ & Definition \ref{def: r-order}\\
            & $\Delta(\G)$ & Maximum degree of MG $\G$& \\
            & $\Delta_{in}(\G)$ & Maximum in-degree of DAG $\G$& \\
            & $\Delta_{in}^+(\G)$ & Maximum size of $\PaP{.}{\G}$ in MG $\G$& Definition \ref{def: pap}\\
            & $\omega(\G)$ & Clique number of MG $\G$& \\
            & $\alpha(\G)$ & Maximum Markov boundary size of MAG $\G$& Definition \ref{def: Mb 2}\\
            & $\G[\Wb]$ & Induced subgraph of MG $\G$ over $\Wb$ & Definition \ref{def: induced subgraph} \\
            & $\G_{\Wb}$ & Latent projection of MAG $\G$ over $\Wb$ & Definition \ref{def: latent projection}\\
            & $[\G]$ & Markov equivalent MAGs of MAG $\G$ & Definition \ref{def: MEC} \\
            & $[\G]^d$ & Markov equivalent DAGs of DAG $\G$ & Definition \ref{def: MEC} \\
            & $\sep{X}{Y}{\Zb}{\G}$ & An m-separation in MG $\G$ & Definition \ref{def: m-separation}\\
            & $\sep{X}{Y}{\Zb}{P}$ & A conditional independence (CI) in distribution $P$ & \\
            & $\D(\Wb)$ & A collection of i.i.d. samples from distribution $P_{\Wb}$ & \\
        \bottomrule
        \end{tabular}
    \end{adjustbox}
    \caption{Table of notations.}
    \label{table of notations}
\end{table}

\section{Preliminaries} \label{sec: preliminaries}
    The key notations are summarized in Table \ref{table of notations} to enhance clarity in our presentation.
    Throughout the paper, we denote random variables in capital letters, sets of variables in bold letters, and graphs in calligraphic letters (e.g., $\G$).
    Further, since the graphs are defined over a set of random variables, we use the terms variable and vertex interchangeably.

    \subsection{Preliminary Graph Definitions}
        A \emph{mixed graph} (MG) is a graph $\G = \langle \Wb,\Eb_1,\Eb_2 \rangle$, where $\Wb$ is a set of vertices, $\Eb_1$ is a set of directed edges, i.e., $\Eb_1 \subseteq \{(X,Y)\mid X,Y \in \Wb \}$,
        and $\Eb_2$ is a set of bi-directed edges, i.e., $\Eb_2 \subseteq \big\{\{X,Y\}\mid X,Y \in \Wb \big\}$.
        All graphs in this paper are assumed to be mixed graphs.
        If a directed edge $X\to Y$ exists in the graph, we say $X$ is a \emph{parent} of $Y$ and $Y$ is a \emph{child} of $X$.
        Two variables are as \emph{neighbors} if a directed or bidirected edge exists between them.
        The \emph{skeleton} of $\G$ is an undirected graph with the same set of vertices $\Wb$ and an edge between $X$ and $Y$ if they are neighbors in $\G$.
        The clique number of $\G$, denoted by $\omega(\G)$, is the size of the largest clique (complete subgraph) of the skeleton of $\G$.
        Further, we denote by $\Delta(\G)$ the maximum degree of a graph $\G$.

        \begin{definition}[{$\G[\Wb]$}] \label{def: induced subgraph}
            For an MG $\G= \langle\Wb', \Eb_1', \Eb_2'\rangle$ and subset $\Wb \subseteq \Wb'$, MG $\G[\Wb]= \langle\Wb, \Eb_1, \Eb_2\rangle$ denotes the induced subgraph of $\G$ over $\Wb$, that is $\Eb_1 = \{(X,Y)\in \Eb_1' \mid X,Y \in \Wb\}$ and $\Eb_2 = \big\{\{X,Y\}\in \Eb_2' \mid X,Y \in \Wb\big\}$.
        \end{definition}
        
        A path $\mathcal{P} = (X_1, \dots, X_k)$ is called \emph{directed} if $X_i \to X_{i+1}$ for all $1\leq i<k$.
        If a directed path exists from $X$ to $Y$, we say $X$ is an \emph{ancestor} of $Y$ (we assume each variable is an ancestor of itself).
        We denote by $\Pa{X}{\G}$, $\Ch{X}{\G}$, $\Ne{X}{\G}$, and $\Anc{X}{\G}$ the set of parents, children, neighbors, and ancestors of $X$ in graph $\G$, respectively.
        Further, for a set of vertices $\Xb$, we define
        \begin{equation*}
            \Anc{\Xb}{\G} \coloneqq \bigcup_{X \in \Xb}\Anc{X}{\G}, \quad
            \Pa{\Xb}{\G} \coloneqq \bigcup_{X \in \Xb}\Pa{X}{\G}.
        \end{equation*}

        \begin{definition}[$\Pi(\Wb)$] \label{def: order}

            For a set $\Wb = \{W_1, W_2, \ldots, W_m\}$, an order over $\Wb$ is a tuple  $(W_{i_1}, W_{i_2}, \ldots, W_{i_n})$, where $\{i_1, i_2, \ldots, i_m\}$ is a permutation of $\{1, 2, \ldots, m\}$.
            We denote by $\Pi(\Wb)$, the set of all orders over $\Wb$.
        \end{definition}
        
        \begin{definition}[$\PaP{X}{\G}$] \label{def: pap}
            The district set of a variable $X$ in MG $\G$, denoted by $\Dis{X}{\G}$, is the set of variables with a path to $X$ comprised only of bi-directed edges.
            By $\PaP{X}{\G}$, we denote the union of parents, district set, and parents of district set.
            That is, 
            \begin{equation*}
                \PaP{X}{\G} \coloneqq \Pa{X}{\G} \cup \Dis{X}{\G} \cup \Pa{\Dis{X}{\G}}{\G}.
            \end{equation*}
            Furthermore, by $\Delta_{in}^+(\G)$, we denote the maximum size of $\PaP{.}{\G}$ in $\G$.
        \end{definition}
        A non-endpoint vertex $X_i$ on a path $(X_1, X_2, \cdots, X_k)$ is called a \emph{collider} if one of the following situations arises.
        \begin{align*}
            X_{i-1} \to X_i \gets X_{i+1},\quad X_{i-1} \leftrightarrow X_i \gets X_{i+1},\\ 
            X_{i-1} \to X_i \leftrightarrow X_{i+1}, \quad
            X_{i-1} \leftrightarrow X_i \leftrightarrow X_{i+1}.
        \end{align*}
        When $X_{i-1}$ and $X_{i+1}$ are not neighbors, and $X_i$ is a collider, the arrangement is termed as an unshielded collider.
        A path $\mathcal{P}$ is a \emph{collider path} if every non-endpoint vertex on $\mathcal{P}$ is a collider on $\mathcal{P}$.

        \begin{definition}[m-separation] \label{def: m-separation}
            In an MG $\G$ over $\Wb$, a path $\mathcal{P}= (X, W_1, \cdots, W_k, Y)$ between two distinct variables $X$ and $Y$ is said to be blocked by a set $\Zb \subseteq \Wb \setminus \{X,Y\}$ in $\G$ if there exists $1 \leq i \leq k$ such that 
            \begin{enumerate*}[label=(\roman*)]
                \item $W_i$ is a collider on $\mathcal{P}$ and $W_i \notin \Anc{\Zb \cup \{X,Y\}}{\G}$, or
                \item $W_i$ is not a collider on $\mathcal{P}$ and $W_i \in \Zb$.
            \end{enumerate*}
            We say $\Zb$ m-separates $X$ and $Y$ in $\G$ and denote it by $\sep{X}{Y}{\Zb}{\G}$ if all the paths in $\G$ between $X$ and $Y$ are blocked by $\Zb$.
        \end{definition}
        A \emph{directed cycle} exists in an MG $\G= \langle\Wb,\Eb_1,\Eb_2\rangle$ when there exists $X,Y \in \Wb$ such that $(X,Y) \in \Eb_1$ and $Y\in \Anc{X}{\G}$.
        Similarly, an \emph{almost directed cycle} exists in $\G$ when there exists $X,Y \in \Wb$ such that $\{X,Y\} \in \Eb_2$ and $Y\in \Anc{X}{\G}$.
        An MG with no directed or almost-directed cycles is said to be \emph{ancestral}.
        An ancestral MG is called \emph{maximal} if every pair of non-neighbor vertices are m-separable, i.e., there exists a set of vertices that m-separates them.
        An MG is called a \emph{maximal ancestral graph} (MAG) if it is both ancestral and maximal.
        A MAG with no bidirected edges is called a \emph{directed acyclic graph (DAG)}.
        When $\G$ is a DAG, the definition of m-separation reduces to the so-called d-separation.
        Moreover, unshielded colliders in a DAG reduce to \emph{v-structures}.
        For a DAG $\G$, we denote its maximum in-degree by $\Delta_{in}(\G)$.

        \begin{definition}[$\VS{X}{\G}$] \label{def: VS}
            In a DAG $\G$, we denote by $\VS{X}{\G}$ the set of v-structures in which vertex $X$ is one of the two parents.
        \end{definition}

        \begin{definition}[Co-parent] \label{def: CP}
		In a DAG, two non-neighbor vertices are co-parents if they have at least one common child.
            The set of co-parents of vertex $X$ in DAG $\G$ is denoted by $\CP{X}{\G}$.
	\end{definition}

        \begin{definition}[Discriminating path] \label{def:discpath}
            In a MAG $\G$, a path $\mathcal{P} = (X, V_1,\dots, V_k, Y)$, where $k \geq 1$, is a discriminating path for $V_k$ if (i) $X$ and $Y$ are not neighbors, (ii) $\{V_1,\dots V_{k-1}\}\subseteq\Pa{X}{\G}$, and (iii) $\{V_1,\dots V_{k-1}\}$ are colliders on $\mathcal{P}$.
        \end{definition}

        \begin{definition}[Inducing path] \label{def: inducing path}
            Suppose $\G$ is a MAG over $\Wb_1 \sqcup \Wb_2$ and let $X, Y$ be distinct vertices in $\Wb_1$.
            An inducing path between $X$ and $Y$ relative to $\Wb_2$ is a path on which (i) every non-collider is a member of $\Wb_2$, and (ii) every collider belongs to $\Anc{X, Y}{\G}$.
        \end{definition}
        We note that in Definition \ref{def: inducing path}, no subset of $\Wb_1$ can block an inducing path relative to $\Wb_2$.
        \begin{definition}[Latent projection] \label{def: latent projection}
            Suppose $\G$ is a MAG over $\Wb_1 \sqcup \Wb_2$.
            The latent projection of $\G$ over $\Wb_1$, denoted by $\G_{\Wb_1}$, is a MAG over $\Wb_1$ constructed as follows:
            \begin{enumerate}[label=(\roman*)]
                \item 
                    Skeleton: $X, Y\in \Wb_1$ are neighbors in $\G_{\Wb_1}$ if there exists an inducing path in $\G$ between $X$ and $Y$ relative to $\Wb_2$.
                \item
                    Orientation: For each pair of neighbors $X,Y$ in $\G_{\Wb_1}$, the edge between $X$ and $Y$ is oriented as $X\to Y$ if $X \in \Anc{Y}{\G}$ and $Y \notin \Anc{X}{\G}$ and as $X \leftrightarrow Y$ if $X \notin \Anc{Y}{\G}$ and $Y \notin \Anc{X}{\G}$.
            \end{enumerate}
        \end{definition}
        \begin{remark}
            The latent projection maintains the ancestral relationships.
            Furthermore, in a MAG $\G$ over $\Wb$, for any $\Wb_2 \subseteq \Wb_1 \subseteq \Wb$ we have $(\G_{\Wb_1})_{\Wb_2} = \G_{\Wb_2}$.
        \end{remark}
        \cite{richardson2002ancestral} showed that the latent projection in Definition \ref{def: latent projection} is the unique projection of a MAG $\G = \langle \Wb_1 \sqcup \Wb_2, \Eb_1, \Eb_2 \rangle$  over $\Wb_1$ that satisfies the following:
        for any distinct variables $X,Y$ in $\Wb_1$ and $\Zb \subseteq \Wb_1 \setminus \{X,Y\}$,
        \begin{equation} \label{eq: msep iff dsep}
            \sep{X}{Y}{\Zb}{\G_{\Wb_1}} 
            \iff
            \sep{X}{Y}{\Zb}{\G}.
        \end{equation}
    
    \subsection{Generative Model: Structural Equation Model (SEM)}
        Consider a DAG $\G$ over $\Vb \sqcup \Ub$, where $\Vb$ and $\Ub$ denote sets of observed and unobserved variables, respectively.
        In a \emph{structural equation model} (SEM) with \emph{causal graph} $\G$, each variable $X \in \Vb \cup \Ub$ is generated as $X = f_X(\Pa{X}{\G},\epsilon_X)$, where $f_X$ is a deterministic function and $\epsilon_X$ is the exogenous variable corresponding to $X$ with an additional assumption that the exogenous variables are jointly independent \citep{pearl2009causality}.
        Such a SEM induces a unique joint distribution $P_{\Vb \cup \Ub}$ over all the variables, which satisfies Markov factorization property with respect to $\G$.
        That is,
        \begin{equation*}
            P_{\Vb \cup \Ub}(\Vb, \Ub) = \prod_{X \in \Vb \cup \Ub} P_{\Vb \cup \Ub}(X \vert \Pa{X}{\G}).
        \end{equation*}
        The marginalized distribution $\PV \coloneqq \sum_{\Ub} P_{\Vb \cup \Ub}$ is the \emph{observational distribution} of the underlying SEM.
        Furthermore, MAG $\GV$ (i.e., the latent projection of DAG $\G$ over $\Vb$ as introduced in Definition \ref{def: latent projection}) is the causal MAG over the observed variables.
        If all the variables in the SEM are observable, i.e., when $\Ub = \varnothing$, $\GV$ coincides with $\G$.
        This assumption is commonly referred to as \emph{causal sufficiency}.
        Therefore, the causal graph over the observed variables is a DAG when causal sufficiency holds.
        
        Next, we show that under suitable assumptions, causal MAG $\GV$ captures the conditional independencies of $\PV$.

    \subsection{Markov Property and Faithfulness} \label{sec: faithfulness}
        Let $P$ be the joint distribution of a set of variables $\Wb$.
        For distinct variables $X, Y \in \Wb$ and $\Zb \subseteq \Wb \setminus \{X,Y\}$, a \emph{conditional independence (CI)} test in $P$ on the triplet $(X, Y, \Zb)$ yields independence, denoted by $\sep{X}{Y}{\Zb}{P}$, if $P(X\vert Y,\Zb) = P(X \vert \Zb)$.
        Distribution $P$ satisfies \emph{Markov property} with respect to MG $\G$ if m-separations in $\G$ imply CIs in $P$.
        That is,
        \begin{equation*}
            \sep{X}{Y}{\Zb}{\G} \Longrightarrow \sep{X}{Y}{\Zb}{P}.
        \end{equation*}
        Conversely, distribution $P$ satisfies \emph{faithfulness} with respect to MG $\G$ if CIs in $P$ imply m-separations in $\G$.
        That is,
        \begin{equation*}
            \sep{X}{Y}{\Zb}{P} \Longrightarrow \sep{X}{Y}{\Zb}{\G}.
        \end{equation*}
        Consider a SEM over $\Vb \sqcup \Ub$ with causal DAG $\G$ and joint distribution $P_{\Vb \cup \Ub}$.
        The joint distribution $P_{\Vb \cup \Ub}$ satisfies Markov property with respect to DAG $\G$ \citep{pearl2000models}.
        Although $P_{\Vb \cup \Ub}$ does not necessarily satisfy faithfulness with respect to $\G$, it is a common assumption in the literature.
        When faithfulness holds, we have
        \begin{equation} \label{eq: faithfulness}
            \sep{X}{Y}{\Zb}{P_{\Vb \cup \Ub}} \iff \sep{X}{Y}{\Zb}{\G},
        \end{equation}
        for any distinct variables $X,Y$ in $\Vb \cup \Ub$ and $\Zb \subseteq \Vb \cup \Ub \setminus \{X,Y\}$.

        Recall that $\Ub$ is the set of unobserved variables.
        Causal discovery aims to find a graphical model over $\Vb$ that encodes the CI relations in $\PV$.
        Note that the induced subgraph $\G[\Vb]$ does not necessarily satisfy Markov property and faithfulness with respect to $\PV$, failing to encode the CIs of $\PV$.
        On the other hand, Equations \eqref{eq: msep iff dsep} and \eqref{eq: faithfulness} imply that the latent projection of $\G$ over $\Vb$, i.e., $\GV$, satisfies Markov property and faithfulness with respect to $\PV$.
        That is, for distinct variables $X,Y$ in $\Vb$ and $\Zb \subseteq \Vb \setminus \{X,Y\}$,
        \begin{equation} \label{eq: PV iff GV}
            \sep{X}{Y}{\Zb}{\PV} \iff \sep{X}{Y}{\Zb}{\GV}.
        \end{equation}
        Accordingly, causal discovery aims to learn $\GV$.
        However, as we discuss next, using the observational distribution $\PV$, MAG $\GV$ can only be learned up to an equivalency class.

    \subsection{Causal Discovery from Observational Distribution} \label{sec: CD from observation}
        By performing CI tests in the observational distribution $\PV$, the m-separations of causal MAG $\GV$ can be recovered using Equation \eqref{eq: PV iff GV}.
        However, two MAGs might have the same set of m-separations.
        
        \begin{definition}[MEC] \label{def: MEC}
            Two MAGs $\G_1$ and $\G_2$ are \emph{Markov equivalent} if they impose the same set of m-separations, i.e., $\sep{X}{Y}{\Zb}{\G_1} \iff \sep{X}{Y}{\Zb}{\G_2}$.
            We denote by $[\G]$ the set of Markov equivalent MAGs of $\G$, known as the Markov equivalence class (MEC).
            Moreover, if $\G$ is a DAG, we denote by $[\G]^d$ the set of Markov equivalent DAGs of $\G$.
        \end{definition}
        CIs are statistically sufficient for causal discovery in non-parametric models\footnote{Side information about the underlying SEM can render the causal graph uniquely identifiable. For instance, when the functions are linear and the exogenous noises are non-Gaussian, CIs are no longer statistically sufficient, and the causal MAG is uniquely identifiable from observational distribution.}.
        That is, if multiple MAGs satisfy Markov property and faithfulness with respect to the observational distribution $\PV$, we cannot identify which one is the causal MAG of the underlying SEM.
        Nevertheless, without the assumption of causal sufficiency, causal discovery from observational distribution aims to identify $[\GV]$.
        When causal sufficiency holds, i.e., $\Ub = \varnothing$ and $\GV = \G$, the goal is to identify $[\G]^d$.

    \subsection{Markov Boundary}
        The majority of the approaches that we present in this paper employ the concept of Markov boundary.
        This notion can be defined based on either a distribution or a graph.

        \begin{definition}[$\Mb{X}{\Wb}$] \label{def: Mb 1}
            Suppose $P$ is the joint distribution of a set of random variables $\Wb$ and let $X \in \Wb$.
            Markov boundary of $X$ with respect to $\Wb$, denoted by $\Mb{X}{\Wb}$, is a minimal subset of $\Wb \setminus \{X\}$, such that 
            \begin{equation*}
                \sep{X}{\Wb \setminus (\Mb{X}{\Wb} \cup \{X\})}{\Mb{X}{\Wb}}{P}.
            \end{equation*}
            Additionally, we define $\MB{\Wb}$ as the set of Markov boundaries of all variables in $\Wb$ with respect to $\Wb$, i.e., $\MB{\Wb} \coloneqq \{\Mb{X}{\Wb} \vert X \in \Wb\}$.
        \end{definition}
        \begin{definition}[$\Mb{X}{\G}$] \label{def: Mb 2}
            In an MG $\G$, the Markov boundary of a variable $X$, denoted by $\Mb{X}{\G}$, consists of all the variables that have a collider path to $X$.
            We denote by $\alpha(\G)$ the maximum Markov boundary size of the variables in $\G$.
        \end{definition}
        \begin{remark}
            If $\G$ is a DAG, then
            \begin{equation*}
                \Mb{X}{\G}
                = \Pa{X}{\G} \cup \Ch{X}{\G} \cup \CP{X}{\G} 
                = \Ne{X}{G} \cup \CP{X}{\G}.
            \end{equation*}
        \end{remark}
        Consider a MAG $\G$ and a distribution $P$, both over a set $\Wb$.
        \cite{pellet2008finding} and \cite{yu2018mining} showed that if $\G$ satisfies Markov property and faithfulness with respect to $P$, then for each variable $X \in \Wb$, $\Mb{X}{\Wb}$ is unique and is equal to $\Mb{X}{\G}$.

\section{Theoretical Foundation of Proposed Algorithms} \label{sec: theory}
    In this section, we establish the theoretical basis for our proposed algorithms in causal discovery.
    Firstly, we present a generic recursive framework for causal discovery in Section \ref{subsec: recursive framework}.
    Then, we explore the concept and characteristics of removable variables in Section \ref{subsec: removable variable}.
    Finally, we compare and contrast the newly proposed removable orders with traditional approaches in Section \ref{subsec: removable orders}.
    
    \subsection{A Recursive Framework for Causal Discovery} \label{subsec: recursive framework}
        Herein, we present a generic recursive framework for causal discovery that does not necessarily assume causal sufficiency.
        Consider a SEM over $\Vb \sqcup \Ub$ with causal DAG $\G$, where $\Vb$ and $\Ub$ denote the set of observed and unobserved variables, respectively.
        As discussed in Section \ref{sec: CD from observation}, our goal is to learn $[\GV]$ (or $[\G]^d$ with the assumption of causal sufficiency) from $\PV$ when Equation \eqref{eq: PV iff GV} holds.
        In the following sections, we mainly focus on recovering the skeleton of $\GV$.
        Later in Section \ref{sec: MEC}, we show that a slight variation of the discussed methods can be employed to recover the MEC.
    
        \begin{definition}[$\Ne{X}{\Wb}$] \label{def: Neighbors}
            Suppose $P_{\Wb}$ is the joint distribution of variables in a set $\Wb$. 
            For $X\in \Wb$, we denote by $\Ne{X}{\Wb}$ the set of variables $Y \in \Wb \setminus \{X\}$ such that for each $\Zb \subseteq \Wb \setminus \{X,Y\}$, we have $\nsep{X}{Y}{\Zb}{P_{\Wb}}$.
        \end{definition}
        Since non-neighbors in any MAG are m-separable, for any $\Wb \subseteq \Vb$ we have
        \begin{equation} \label{eq: Neighbors in graph and distribution}
            \Ne{X}{\Wb} = \Ne{X}{\G_{\Wb}}.
        \end{equation}
        Hence, to learn the skeleton of $\GV$, it suffices to learn $\Ne{X}{\Vb}$ for each $X \in \Vb$.
        In practice, a finite set of samples from $\PV$ is available instead of knowing the exact probability distribution.
        Let $\D(\Vb)$ denote a collection of i.i.d. samples from $\PV$, sufficiently large to recover the CI relations in $\PV$.
        For any subset $\Wb \subseteq \Vb$, $\D(\Wb)$ represents the samples corresponding to the variables in $\Wb$.

        \begin{algorithm}[h!]
            \caption{Recursive framework for learning the skeleton of $\GV$}
            \label{alg: Recursive framework}
            \begin{algorithmic}[1]
                \STATE \textbf{Input:} $\D(\Vb)$
                \STATE $\Vb_1 \gets \Vb$
                \FOR{$i$ from $1$ to $n-1$}
                    \STATE $X_i \gets $ Find a [\textit{removable}] variable in $\G_{\Vb_i}$ using $\D(\Vb_i)$
                    \STATE Learn $\Ne{X_i}{\Vb_i}$ using $\D(\Vb_i)$
                    \STATE $\Vb_{i+1} \gets \Vb_i \setminus \{X_i\}$
                \ENDFOR
            \end{algorithmic}
        \end{algorithm}
        
        A generic recursive framework for causal discovery is presented in Algorithm \ref{alg: Recursive framework}.
        The algorithm iteratively removes variables from $\Vb$ and learns the skeleton over the remaining variables.
        At the $i$-th iteration, a variable $X_i \in \Vb_i$ is selected, where $\Vb_i$ denotes the set of remaining variables.
        As we shall discuss shortly, this variable cannot be arbitrary and must be \textit{removable}.
        After finding such a variable $X_i$, the algorithm learns the set of neighbors of $X_i$ in $\G_{\Vb_i}$ using $\D(\Vb_i)$.
        Note that this is possible because of Equation \eqref{eq: Neighbors in graph and distribution}.
        Then, the samples corresponding to $X_i$ are discarded, and the causal discovery problem is solved recursively for the remaining variables $\Vb_{i+1} =\Vb_i \setminus \{X_i\}$.
    
        For Algorithm \ref{alg: Recursive framework} to correctly learn the skeleton of $\GV$, it is necessary that at each iteration, $\Ne{X_i}{\Vb_i} = \Ne{X_i}{\GV[\Vb_i]}$.
        Recall that $\Ne{X_i}{\Vb_i} = \Ne{X_i}{\G_{\Vb_i}}$ and $\GV[\Vb_i]$ denotes the induced subgraph of $\GV$ over $\Vb_i$.
        Since the latent projection can only add new edges to the projected graph, this condition is equivalent to
        \begin{equation} \label{eq: induced = projection}
            \G_{\Vb_i} = \GV[\Vb_i], \quad \forall 1\leq i < n.
        \end{equation}
        However, selecting an arbitrary variable $X_i$ in line $4$ may not uphold this property.
    
        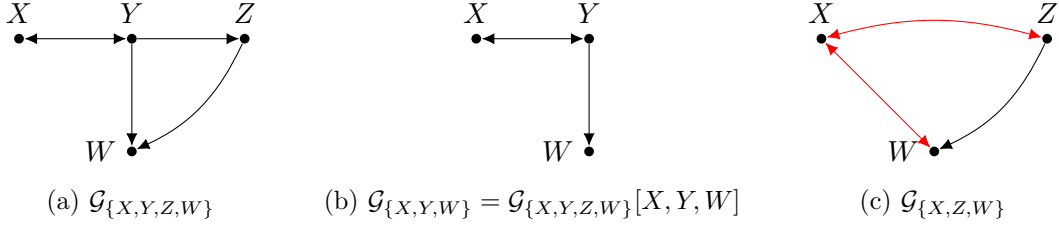
\begin{figure}[t!] 
            \centering
            \tikzstyle{block} = [circle, inner sep=1.3pt, fill=black]
            \tikzstyle{input} = [coordinate]
            \tikzstyle{output} = [coordinate]
            \tikzset{edge/.style = {->,> = latex',-{Latex[width=1.5mm]}}}
            \tikzset{bedge/.style={<->, > = latex', {Latex[width=1.5mm]}-{Latex[width=1.5mm]}}}
    
            \begin{subfigure}[b]{0.3\textwidth}
                \centering   
                \begin{tikzpicture}
                    \node[block] (X) at  (0,0) {};
                    \node[] ()[above=0 of X]{$X$};
                    \node[block] (Y) at  (1.5,0) {};
                    \node[] ()[above=0 of Y]{$Y$};
                    \node[block] (Z) at  (3,0) {};
                    \node[] ()[above=0 of Z]{$Z$};
                    \node[block] (W) at  (1.5,-1.5) {};
                    \node[] ()[left=0 of W]{$W$};
                    \draw[bedge] (X) to (Y);
                    \draw[edge] (Y) to (Z);
                    \draw[edge] (Y) to (W);
                    \draw[edge, bend left=20] (Z) to (W);
                \end{tikzpicture}
                \caption{$\G_{\{X, Y, Z, W\}}$}
                \label{fig: a.rem}
            \end{subfigure} \hfill
            \begin{subfigure}[b]{0.38\textwidth}
                \centering
                \begin{tikzpicture}
                    \node[block] (X) at  (0,0) {};
                    \node[] ()[above=0 of X]{$X$};
                    \node[block] (Y) at  (1.5,0) {};
                    \node[] ()[above=0 of Y]{$Y$};
                    \node[block] (W) at  (1.5,-1.5) {};
                    \node[] ()[left=0 of W]{$W$};
                    \draw[bedge] (X) to (Y);
                    \draw[edge] (Y) to (W);
                \end{tikzpicture}
                \caption{$\G_{\{X, Y, W\}} = \G_{\{X, Y, Z, W\}}[X, Y, W]$}
                \label{fig: b.rem}
            \end{subfigure} \hfill
            \begin{subfigure}[b]{0.3\textwidth}
                \centering
                \begin{tikzpicture}
                    \node[block] (X) at  (0,0) {};
                    \node[] ()[above=0 of X]{$X$};
                    \node[block] (Z) at  (3,0) {};
                    \node[] ()[above=0 of Z]{$Z$};
                    \node[block] (W) at  (1.5,-1.5) {};
                    \node[] ()[left=0 of W]{$W$};
                    \draw[edge, bend left=20] (Z) to (W);
                    \draw[bedge, red, bend left=15] (X) to (Z);
                    \draw[bedge, red] (X) to (W);
                \end{tikzpicture}
                \caption{$\G_{\{X, Z, W\}}$}
                \label{fig: c.rem}
            \end{subfigure}
            \caption{Graphs in Example \ref{example: must remove removable}.}
            \label{fig: rem}
        \end{figure}
    
        \begin{example} \label{example: must remove removable}
            Consider MAG $\G_{\{X,Y,Z,W\}}$ shown in Figure \ref{fig: a.rem}.
            In Figure \ref{fig: b.rem}, MAG $\G_{\{X,Y,W\}}$ is the same as $\G_{\{X, Y, Z, W\}}[X,Y,W]$. However, in Figure \ref{fig: c.rem}, MAG $\G_{\{X,Z,W\}}$ has two extra edges between $X,Z$ and $X,W$, which are not present in $\G_{\{X, Y, Z, W\}}[X,Z,W]$.
            This demonstrates that when $\Vb_i = \{X, Y, Z, W\}$ in Algorithm \ref{alg: Recursive framework}, $Z$ can be selected in line $4$, whereas $Y$ cannot.
        \end{example}
        To employ Algorithm \ref{alg: Recursive framework}, we need to provide a method to find a removable variable and learn its neighbors at each iteration.
        In the next section, we will define the concept of removable variables and show that Equation \eqref{eq: induced = projection} holds \emph{if and only if} a removable variable is selected at each iteration of this recursive framework.
    
    \subsection{Removable Variables} \label{subsec: removable variable}
        In this section, we define removable variables, present a graphical characterization for them under different assumptions, and provide certain crucial properties.
        
        \begin{definition}[Removable variable]\label{def: removable}
            In a MAG $\G$ over $\Wb$, a vertex $X \in \Wb$ is called removable if $\G$ and $\G[\Wb \setminus \{X\}]$ impose the same set of m-separations over $\Wb \setminus \{X\}$. 
            That is, for any distinct vertices $Y,T\in \Wb \setminus \{X\}$ and $\Zb \subseteq \Wb \setminus \{X, Y, T\}$,
            \begin{equation*}
                \sep{Y}{T}{\Zb}{\G}
                \iff
                \sep{Y}{T}{\Zb}{\G[\Wb \setminus \{X\}]}.
            \end{equation*}
        \end{definition}
        
        \begin{proposition}[Only removables can get removed] \label{prp: removable}
            Suppose $\G$ is a MAG over $\Wb$ and $X \in \Wb$.
            MAG $\G_{\Wb \setminus \{X\}}$ is equal to $\G[\Wb \setminus \{X\}]$ if and only if $X$ is removable in $\G$.
        \end{proposition}
        Proposition \ref{prp: removable} implies that Equation \eqref{eq: induced = projection} holds for Algorithm \ref{alg: Recursive framework} if and only if $X_i$ is removable in $\G_{\Vb_i}$ at each iteration.
        Next, we provide graphical characterizations of removable variables in both MAGs and DAGs, along with their key properties.
        We then define removable orders and discuss their properties.

        \subsubsection{Graphical Characterization of Removable Variables}
            We present graphical characterizations of removable variables in DAGs and MAGs.
        
            \begin{figure}[t!] 
                \centering
                \tikzstyle{block} = [circle, inner sep=1.3pt, fill=black]
                \tikzstyle{input} = [coordinate]
                \tikzstyle{output} = [coordinate]
                \begin{subfigure}[b]{0.49\textwidth}
                    \centering
                    \begin{tikzpicture}
                        \tikzset{edge/.style = {->,> = latex',-{Latex[width=1.5mm]}}}
                        \node [block](X) {};
                        \node[] ()[above =0 of X]{$X$};
                        \node [block, below right= 1.2cm and 1.2cm of X](Z) {};
                        \node[] ()[below left=-0.1cm and -0.12cm of Z]{$Z$};
                        \node [block, right= 1.2cm of X](Y) {};
                        \node[] ()[above =0 of Y]{$Y$};
                        \draw[edge] (X) to (Z);
                        \draw[edge] (Y) to (X);
                        \draw [edge, line width=0.5mm, red] (Y) to (Z);
                    \end{tikzpicture}
                    \hfill
                    \begin{tikzpicture}
                        \tikzset{edge/.style = {->,> = latex',-{Latex[width=1.5mm]}}}
                        \node [block](X) {};
                        \node[] ()[above =0 of X]{$X$};
                        \node [block, below right= 1.2cm and 1.2cm of X](Z) {};
                        \node[] ()[below left=-0.1cm and -0.12cm of Z]{$Z$};
                        \node [block, right= 1.2cm of X](Y) {};
                        \node[] ()[above =0 of Y]{$Y$};
                        \node[color=red,style={minimum size=0.1cm},inner sep=0.01pt](stary) [below= 0 of Y]{$*$};
                        \node[color=red,style={minimum size=0.1cm},inner sep=0.01pt](starz) [above= 0 of Z]{$*$};
                        \draw[edge] (X) to (Z);
                        \draw[edge] (X) to (Y);
                        \draw [edge, -, line width=0.5mm, red] (stary) to (starz);
                    \end{tikzpicture}
                    \hfill
                    \begin{tikzpicture}
                        \tikzset{edge/.style = {->,> = latex',-{Latex[width=1.5mm]}}}
                        \node [block](X) {};
                        \node[] ()[above =0 of X]{$X$};
                        \node [block, below right= 1.2cm and 1.2cm of X](Z) {};
                        \node[] ()[below left=-0.1cm and -0.12cm of Z]{$Z$};
                        \node [block, right = 1.2cm of X](V) {};
                        \node[] ()[above =0 of V]{$V$};
                        \node [block, right= 1.2cm of V](Y) {};
                        \node[] ()[above =0 of Y]{$Y$};
                        \draw[edge] (X) to (V);
                        \draw[edge] (X) to (Z);
                        \draw[edge] (V) to (Z);
                        \draw[edge] (Y) to (V);
                        \draw [edge, line width=0.5mm, red] (Y) to (Z);
                    \end{tikzpicture}
                    \caption{For DAGs}
                    \label{fig: graph-rep a}
                \end{subfigure}
                \hfill
                \begin{subfigure}[b]{0.49\textwidth}
                    \centering
                    \begin{tikzpicture}
                        \tikzset{edge/.style = {->,> = latex',-{Latex[width=1.5mm]}}}
                        \node[block] (x) at  (0,0) {};
                        \node[] ()[above =0 of x]{$X$};
                        \node[block] (v1) at  (1.5,0) {};
                        \node[] ()[above =0 of v1]{$V_1$};
                        \node[style={minimum size=0.1cm},inner sep=0.01pt](starv1tx) [right= 0 of x]{$*$};
                        \node[block] (v2) at  (3,0) {};
                        \node[] ()[above =0 of v2]{$V_2$};
                        \node[right=0.4cm of v2]{\ldots};
                        \node[block] (vm) at  (4.5,0) {};
                        \node[] ()[above =0 of vm]{$V_m$};
                        \node[block] (y) at  (6,0) {};
                        \node[] ()[above =0 of y]{$Y$};
                        \node[style={minimum size=0.1cm},inner sep=0.01pt](starvmty) [left= 0 of y]{$*$};
                        \node[block] (z) at  (1.5,-1.5) {};
                        \node[] ()[above left=-0.5cm and -0.1cm of z]{$Z$};
                        \node[color=red,style={minimum size=0.1cm},inner sep=0.01pt](starY) [below left=0.03cm and -0.1cm of y]{$*$};
                        \node[color=red,style={minimum size=0.1cm},inner sep=0.01pt](starZ) [below right =0 and 0 of z]{$*$};
                        \draw[edge] (starv1tx) to (v1);
                        \draw[edge] (v1) to (v2);
                        \draw[edge] (v2) to (v1);
                        \draw[edge] (starvmty) to (vm);
                        \draw[edge] (x) to (z);
                        \draw[edge] (v1) to (z);
                        \draw[edge] (v2) to (z);
                        \draw[edge, bend left = 20] (vm) to (z);
                        \draw[edge, line width=0.5mm, color=red, style={-}, bend left = 25] (starY) to (starZ);
                    \end{tikzpicture}
                    \caption{For MAGs}
                    \label{fig: graph-rep b}
                \end{subfigure}
                \caption{Graphical criterion of removability.
                Asterisk (\raisebox{-0.5ex}{*}) is used as a wildcard, which indicates that the edge endpoint can be either an arrowhead or a tail.
                In the case of MAGs, the path $(X,V_1,...,Y)$ is a collider path and $X,...,V_m\in\Pa{Z}{\G}$. $X$ is removable if and only if for all such paths, $Y$ and $Z$ are adjacent.
                In the case of a DAG, it suffices to check this condition for vertices $Y$, where $Y$ is a parent, child, or co-parent of $X$.}
                \label{fig: graph-rep} 
            \end{figure}
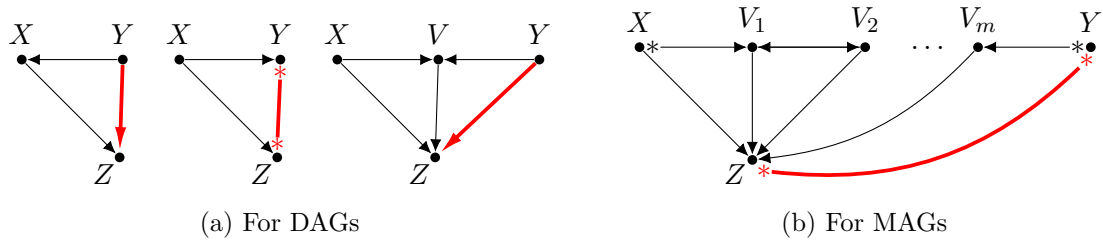

            \begin{theorem}[Graphical characterization in DAGs] \label{thm: graph-rep DAG}
                A variable $X$ is removable in a DAG $\G$ if and only if the following two conditions are satisfied for every $Z \in \Ch{X}{\G}$.
                \begin{enumerate}
                    \item[]
                        \textbf{Condition 1:} $\Ne{X}{\G} \subset \Ne{Z}{\G} \cup \{Z\}$.
                    \item[]
                        \textbf{Condition 2:} $\Pa{V}{\G} \subset \Pa{Z}{\G}$, \quad $\forall V \in \Ch{X}{\G} \cap \Pa{Z}{\G}$.
                \end{enumerate}
            \end{theorem}
            
            Figure \ref{fig: graph-rep a} depicts the two conditions of Theorem \ref{thm: graph-rep DAG} for removable variables in DAGs.
        
            \begin{theorem}[Graphical characterization in MAGs] \label{thm: graph-rep MAG}
                Let $\G$ be a MAG over $\Wb$.
                Vertex $X$ is removable in $\G$ if and only if for any collider path $u=(X,V_1,...,V_m,Y)$ and $Z \in \Wb \setminus \{X,Y,V_1,...,V_m\}$ such that $\{X,V_1,...,V_m\} \subseteq \Pa{Z}{\G}$, $Y$ and $Z$ are neighbors.
            \end{theorem}
            Figure \ref{fig: graph-rep b} depicts the condition of Theorem \ref{thm: graph-rep MAG} for removable variables in MAGs.
            Path $u=(X,V_1,...,V_m,Y)$ is a collider path where $\{X,V_1,...,V_m\}\subseteq\Pa{Z}{\G}$.
            Theorem \ref{thm: graph-rep MAG} states that $X$ is removable if and only if, for any such path, $Y$ and $Z$ are neighbors.
            In the case of a DAG, the collider paths are of size either $1$ (parents and children of $X$) or $2$ (co-parents of $X$).
            As shown in Figure \ref{fig: graph-rep a}, the graphical criterion of Theorem \ref{thm: graph-rep MAG} reduces to checking the adjacency of these three groups of vertices with each child of $X$ (see Theorem \ref{thm: graph-rep DAG}).
    
        \subsubsection{Properties of Removable Variables}
            Herein, we discuss some key properties of removable variables.
            \begin{proposition}[Removables exist] \label{prp: removable exists}
                In any MAG, the variables with no children are removable, and the set of vertices with no children is non-empty.
                Therefore, any MAG has at least one removable variable.
            \end{proposition}
            Proposition \ref{prp: removable exists} implies that $X_i$ in line $4$ of Algorithm \ref{alg: Recursive framework} is well-defined.
            \begin{proposition}[Removables have small Mb size] \label{prp: small Mb}
                In a MAG $\G$, if a vertex $X$ is removable, then $|\Mb{X}{\G}| \leq \Delta_{in}^+(\G)$.
                Furthermore, if $\G$ is a DAG, then $|\Mb{X}{\G}| \leq \Delta_{in}(\G)$.
            \end{proposition}
            Note that for an arbitrary variable $X$, $|\Mb{X}{\G}|$ can be as large as $n$ whereas $\Delta_{in}^+(\G)$ (or $\Delta_{in}(\G)$ for DAGs) is typically a small number, demonstrating that removable variables have relatively small Markov boundary sizes.
            \begin{proposition}[Removables are invariant in a MEC] \label{prp: invariant}
                Two Markov equivalent MAGs have the same set of removable variables.
            \end{proposition}
            We note that the set of vertices without children is not the same for all MAGs in a MEC. 
            However, Proposition \ref{prp: invariant} implies that the set of removable variables is a superset of the vertices without children, which is invariant across all MAGs in a MEC.

    \subsection{Removable Orders} \label{subsec: removable orders}
        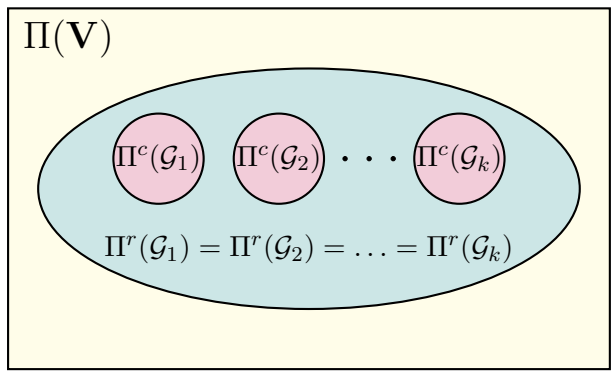
\begin{figure}[b!]
            \centering
            \begin{tikzpicture}[scale=0.8]
                \filldraw[fill=yellow!10, draw=black, thick] (0,0) rectangle (10,6);
                \node at (1,5.5) {\Large $\Pi(\Vb)$};
            
                \filldraw[fill=teal!20, draw=black, thick] (5,3) ellipse (4.5cm and 2cm);
                \node at (5,2) {$\Pi^r(\G_1) = \Pi^r(\G_2) = \ldots = \Pi^r(\G_k)$};
            
                \filldraw[fill=purple!20, draw=black, thick] (2.5,3.5) circle (0.75cm) node[black] {$\Pi^c(\G_1)$};
                \filldraw[fill=purple!20, draw=black, thick] (4.5,3.5) circle (0.75cm) node[black] {$\Pi^c(\G_2)$};
                \filldraw[fill=purple!20, draw=black, thick] (7.5,3.5) circle (0.75cm) node[black] {$\Pi^c(\G_k)$};
            
                \filldraw [black] (5.6,3.5) circle (1pt);
                \filldraw [black] (6,3.5) circle (1pt);
                \filldraw [black] (6.4,3.5) circle (1pt);
            \end{tikzpicture}
            \caption{In this figure, $\{\G_1,\dots,\G_k\}$ denotes a set of Markov equivalent DAGs.
            $\Pi(\Vb)$ denotes the set of orders over $\Vb$, which is the search space of ordering-based methods.
            $\Pi^c(\G_i)$ denotes the set of c-orders of $\G_i$, the target space of existing ordering-based methods in the literature.
            $\Pi^r(\G_i)$ denotes the set of r-orders of $\G_i$, which is the target space of ROL.}
            \label{fig: venn}
        \end{figure}
        
        In Algorithm \ref{alg: Recursive framework}, our generic framework can be viewed as an ordering-based approach.
        In this approach, a variable is eliminated at each iteration based on a specific order.
        In this section, we first define c-orders, which are standard orders that existing ordering-based methods use to recover the graph.
        We then introduce r-order, which are orders that can be integrated into Algorithm \ref{alg: Recursive framework}.
        Finally, we show that r-orders are advantageous over c-orders for structure learning, because of the properties that are depicted in Figure \ref{fig: venn}.
    
        Score-based methods are one of the main classes of algorithms for causal discovery.
        These algorithms use a score function, such as a regularized likelihood function or Bayesian information criterion (BIC), to evaluate graphs and determine the structure that maximizes the score.
        Under the causal sufficiency assumption, the search space of the majority of these methods is the space of DAGs, which contains $2^{\Omega(n^2)}$ members.
        The first ordering-based search strategy was introduced by \citet{teyssier2005ordering}.
        These methods search through the space of orders (Definition \ref{def: order}), which includes $2^{\mathcal{O}(n \log(n))}$ members.
        It is worth noting that the space of orders is much smaller than the space of DAGs.
        Ordering-based methods divide the learning task into two stages.
        In the first stage, they use the available data to find a causal order (c-order), which is defined as follows.
        \begin{definition}[c-order] \label{def: c-order}
            In a DAG $\G = \langle \Wb, \Eb_1, \varnothing \rangle$, an order $\pi = (X_1,\dots, X_m) \in \Pi(\Wb)$ is called causal (in short c-order) if $i>j$ for each $(X_i, X_j) \in \Eb_1$.
            Equivalently, $\pi$ is c-order if $X_i$ has no children in $\G[\{X_i,X_{i+1},\dots, X_n\}]$ for each $1 \leq i \leq n$.
            We denote by $\Pi^c(\G)$ the set of c-orders of $\G$.
        \end{definition}
        We note that c-orders are defined over DAGs, and accordingly, most of the ordering-based approaches for causal discovery require causal sufficiency.
        In the second stage, ordering-based methods use the learned order to identify the MEC of $\G$.
        
        We introduce a novel type of order for MAGs, called removable order (in short, r-order), and argue that r-orders are advantageous over c-orders for structure learning.
        \begin{definition}[r-order] \label{def: r-order}
            In a MAG $\G$, an order $\pi = (X_1,\cdots, X_n)$ is called removable (r-order) if $X_i$ is a removable variable in $\G[\{X_i,,X_{i+1},\dots, X_n\}]$ for each $1 \leq i \leq n$.
            We denote by $\Pi^r(\G)$ the set of r-orders of $\G$.
        \end{definition}
        Note that r-orders are defined over MAGs, which enables us to design ordering-based methods that do not assume causal sufficiency.
        
        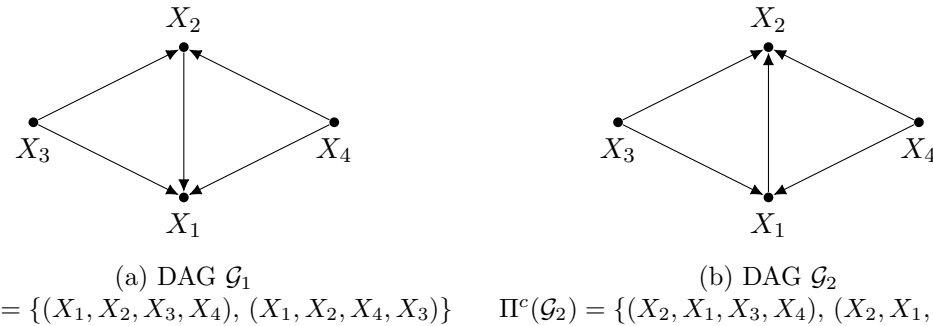
\begin{figure}[t!] 
	    \centering
		\tikzstyle{block} = [circle, inner sep=1.3pt, fill=black]
		\tikzstyle{input} = [coordinate]
		\tikzstyle{output} = [coordinate]
		\begin{subfigure}{0.49\textwidth}
    		\centering
                \begin{tikzpicture}
                    \tikzset{edge/.style = {->,> = latex',-{Latex[width=1.5mm]}}}
                    \node[block] (X1) at  (2,-1) {};
                    \node[] ()[below=0 of X1]{$X_1$};
                    \node[block] (X2) at  (2,1) {};
                    \node[] ()[above=0 of X2]{$X_2$};
                    \node[block] (X3) at  (0,0) {};
                    \node[] ()[below=0 of X3]{$X_3$};
                    \node[block] (X4) at  (4,0) {};
                    \node[] ()[below=0 of X4]{$X_4$};
                    \draw[edge] (X3) to (X2);
                    \draw[edge] (X3) to (X1);
                    \draw[edge] (X4) to (X2);
                    \draw[edge] (X4) to (X1);
                    \draw[edge] (X2) to (X1);
                \end{tikzpicture}
                \captionsetup{justification=centering}
                \caption{DAG $\G_1$\\
                 $\Pi^c(\G_1) = \{(X_1,X_2,X_3,X_4),\, (X_1,X_2,X_4,X_3)\}$}
                \label{fig: c-order and r-order 1}
            \end{subfigure}
            \hfill
            \begin{subfigure}{0.49\textwidth}
        	\centering
                \begin{tikzpicture}
                    \tikzset{edge/.style = {->,> = latex',-{Latex[width=1.5mm]}}}
                    \node[block] (X1) at  (2,-1) {};
                    \node[] ()[below=0 of X1]{$X_1$};
                    \node[block] (X2) at  (2,1) {};
                    \node[] ()[above=0 of X2]{$X_2$};
                    \node[block] (X3) at  (0,0) {};
                    \node[] ()[below=0 of X3]{$X_3$};
                    \node[block] (X4) at  (4,0) {};
                    \node[] ()[below=0 of X4]{$X_4$};
                    \draw[edge] (X3) to (X2);
                    \draw[edge] (X3) to (X1);
                    \draw[edge] (X4) to (X2);
                    \draw[edge] (X4) to (X1);
                    \draw[edge] (X1) to (X2);
                \end{tikzpicture}
                \captionsetup{justification=centering}
                \caption{DAG $\G_2$
                 \\ $\Pi^c(\G_2) = \{(X_2,X_1,X_3,X_4),\,(X_2,X_1,X_4,X_3)\}$}
                \label{fig: c-order and r-order 2}
            \end{subfigure}
            \caption{Two Markov equivalent DAGs $\G_1$ and $\G_2$ that form a MEC together and their disjoint sets of c-orders.
            In this example, any order over $\Vb = \{X_1,X_2,X_3,X_4\}$ is an r-order, i.e., $\Pi^r(\G_1) = \Pi^r(\G_2) = \Pi(\Vb)$. Note that $|\Pi^r(\G_1)| =|\Pi^r(\G_2)|=24 > 2 = |\Pi^c(\G_1)| =|\Pi^c(\G_2)|$.}
            \label{fig: c-order and r-order}
        \end{figure}

        \begin{example}
            Consider the two DAGs $\G_1$ and $\G_2$ and their sets of c-orders depicted in Figure \ref{fig: c-order and r-order}.
            In this case, $\G_1$ and $\G_2$ are Markov equivalent and together form a MEC.
            Furthermore, $\Pi^c(\G_1)$ and $\Pi^c(\G_2)$ are disjoint, and each contains $2$ c-orders, and any order over the set of vertices is an r-order for both $\G_1$ and $\G_2$.
            Hence, each graph has $24$ r-orders.
        \end{example}

        In a MEC, all MAGs share the same r-orders.
        In DAGs, r-orders include all c-orders as subsets. 
        This is illustrated in Figure \ref{fig: venn} and formalized in the subsequent propositions.

        \begin{proposition}[r-orders are invariant across MEC] \label{prp: r-order is learnable}
            If $\G_1$ and $\G_2$ are two Markov equivalent MAGs, then $\Pi^r(\G_1) = \Pi^r(\G_2)$.
        \end{proposition}
        \begin{proposition}[r-orders include c-orders] \label{prp: c-o subset of r-o}
            For any DAG $\G$, we have $\Pi^c(\G) \subseteq \Pi^r(\G)$.
        \end{proposition}

\section{RCD Methods: MARVEL, L-MARVEL, RSL, ROL} \label{sec: RCD methods}
    In the previous section, we introduced a recursive framework for causal discovery.
    We defined removable variables and provided graphical characterizations of them in MAGs and DAGs, along with their important properties.
    Building on that framework, in this section, we present four recursive causal discovery methods: MARVEL, L-MARVEL, RSL, and ROL.
    MARVEL, L-MARVEL, and RSL provide various approaches for lines 4--5 of Algorithm \ref{alg: Recursive framework} under different sets of assumptions.
    ROL, however, identifies all removable variables at once using r-orders.

    \subsection{Finding Removable Variables and Their Neighbors}
        Motivated by Proposition \ref{prp: small Mb}, we provide Algorithm \ref{alg: FindRemovable}, a framework for finding a removable variable.
        \begin{algorithm}
            \caption{Finding a removable variable.}
            \label{alg: FindRemovable}
            \begin{algorithmic}[1]
                \STATE {\bfseries FindRemovable}($\D(\Wb)$, $\MB{\Wb}$)
                \STATE $m \gets |\Wb|$
                \STATE $(W_1, \dots, W_m) \gets$ Sort $\Wb$ such that $|\Mb{W_1}{\Wb}| \leq |\Mb{W_2}{\Wb}|\leq \dots \leq |\Mb{W_m}{\Wb}|$
                \FOR{$i=1$ to $m$}
                    \IF{\textbf{IsRemovable}($W_i$, $\D(\Mb{W_i}{\Wb})$) is \true}
                        \RETURN $W_i$
                    \ENDIF
                \ENDFOR
            \end{algorithmic}
        \end{algorithm}
        
        Given a set $\Wb$, the algorithm takes the Markov boundaries as input, sorts the variables based on the size of their Markov boundaries, and applies the function \textit{IsRemovable} to them.
        The function \textit{IsRemovable} determines whether a variable $W_i$ is removable in $\G_{\Wb}$.
        Algorithm \ref{alg: FindRemovable} returns the first removable variable it identifies and halts.
        Therefore, Proposition \ref{prp: small Mb} implies that \textit{IsRemovable} will only be called for variables with Markov boundary sizes less than or equal to $\Delta_{in}^+(\G)$.
        As we showed in Theorems \ref{thm: graph-rep DAG} and \ref{thm: graph-rep MAG}, the removability of a variable is a property of the causal graph over its Markov boundary.
        Therefore, we only need to utilize data from the variables in $\Mb{W_i}{\Wb}$ for the \textit{IsRemovable} function.
        
        Next, we introduce three approaches for efficiently testing the removability of a variable under different sets of assumptions.

    \subsection{MARVEL: \texorpdfstring{\underline{MA}rkov boundary-based \underline{R}ecursive \underline{V}ariable \underline{EL}imination}{MARVEL: MAarkov boundary-based Recursive Variable ELimination}}

        The first recursive approach that introduced and utilized the notion of removability is MARVEL \citep{mokhtarian2021recursive}.
        MARVEL assumes causal sufficiency, i.e., the causal graph is a DAG.

        Suppose $\G$ is a DAG with the set of vertices $\Vb$.
        To verify the removability of a variable $X\in \Vb$, MARVEL first learns the neighbors of $X$, i.e., $\Ne{X}{\G}$, and the set of v-structures in which $X$ is a parent, i.e., $\VS{X}{\G}$, using the following two lemmas.

        \begin{lemma}[\citealp{pellet2008finding}] \label{lem: neighbor}
            Suppose $\G$ is a MAG over $\Vb$.
            For $X\in \Vb$ and $Y \in \Mb{X}{\G}$, $Y$ is a neighbor of $X$ if and only if 
            \begin{equation} \label{eq: neighbor}
                \nsep{X}{Y}{\Sb}{\G}, \quad \forall \Sb \subsetneq \Mb{X}{\G} \setminus \{Y\}.
            \end{equation}
        \end{lemma}
        We note that while MARVEL uses Lemma \ref{lem: neighbor} when $\G$ is a DAG, \cite{pellet2008finding} showed that it also holds for MAGs.
        
        \begin{lemma} \label{lem: v-structures}
            Suppose $\G$ is a DAG, $Y \in \CP{X}{\G}$, $\sep{X}{Y}{\Sb_{XY}}{\G}$, and $Z \in \Ne{X}{\G}$.
            $Z$ is a common child of $X$ and $Y$, i.e., $(X\to Z \gets Y) \in \VS{X}{\G}$, if and only if
            \begin{equation*}
                Z \notin \Sb_{XY} \quad \text{and} \quad \nsep{Y}{Z}{\Sb}{\G}, \quad \forall \Sb \subseteq \Mb{X}{\G} \cup\{X\} \setminus \{Y,Z\}.
            \end{equation*}
        \end{lemma}
        After learning $\Ne{X}{\G}$ and $\VS{X}{\G}$ using Lemmas \ref{lem: neighbor} and \ref{lem: v-structures}, MARVEL utilizes the following two lemmas to verify the two conditions of Theorem \ref{thm: graph-rep DAG}.
        
        \begin{lemma} \label{lem: condition1}
            Variable $X$ satisfies Condition 1 of Theorem \ref{thm: graph-rep DAG} if and only if
            \begin{equation*}
                \nsep{Y}{Z}{\Sb \cup \{X\}}{\G}, \quad \forall Y, Z \in \Ne{X}{\G},\, \Sb \subseteq \Mb{X}{\G} \setminus \{Y, Z\}.
            \end{equation*}
        \end{lemma}
        
        \begin{lemma} \label{lem: condition2}
            Suppose variable $X$ satisfies Condition 1 of Theorem \ref{thm: graph-rep DAG}.
            Then $X$ satisfies Condition 2 of Theorem \ref{thm: graph-rep DAG}, and therefore, $X$ is removable in $\G$, if and only if
            \begin{equation*}
                \begin{gathered}
                    \nsep{Y}{Z}{\Sb \cup \{X, V\}}{\G},\\
                    \forall (X \to V\gets Y) \in \VS{X}{\G}, Z \in \Ne{X}{\G} \setminus \{V\}, \Sb \subseteq \Mb{X}{\G} \setminus \{V, Y, Z\}.
                \end{gathered}
            \end{equation*}
        \end{lemma}
        The following corollary outlines the computational complexity of applying these lemmas.
        
        \begin{corollary}
            Given $\Mb{X}{\G}$, by applying Lemmas \ref{lem: neighbor}-\ref{lem: condition2}, we can identify $\Ne{X}{\G}, \CP{X}{\G}$, $\VS{X}{\G}$, and determine whether $X$ is removable in $\G$ by performing at most 
            \begin{equation*}
                \mathcal{O}\left(|\Mb{X}{\G}|^2 2^{|\Mb{X}{\G}|}\right)
            \end{equation*} 
            unique CI tests.
        \end{corollary}

    \subsection{L-MARVEL: Latent-MARVEL}
        L-MARVEL extends MARVEL to the case where causal sufficiency does not necessarily hold, i.e., the causal graph is a MAG \citep{akbari2021recursive}.\footnote{L-MARVEL, as presented in \cite{akbari2021recursive}, can also handle the presence of selection bias. In this paper however, for the sake of simplicity, we assume there is no selection bias, and we have access to i.i.d. samples from $P_{\Vb}$.}
        To verify the removability of a variable $X$ in a MAG $\G$, L-MARVEL first learns $\Ne{X}{\G}$ using Lemma \ref{lem: neighbor} as follows.
        If $Y \in \Mb{X}{\G}$ is not a neighbor of $X$, then $X$ and $Y$ have a separating set in $\Mb{X}{\G} \setminus \{Y\}$.
        Hence, identifying $\Ne{X}{\G}$ can be performed using a brute-force search in the Markov boundary, using at most 
        \begin{equation*}
            |\Mb{X}{\G}| 2^{|\Mb{X}{\G}|-1}
        \end{equation*}
        CI tests.
        After learning $\Ne{X}{\G}$, L-MARVEL utilizes the following theorem to check the removability of $X$.
        
        \begin{theorem} \label{thm: test removability} 
            In a MAG $\G$ over $\Vb$, a variable $X \in \Vb$ is removable if and only if for every $Y \in \Mb{X}{\G}$ and $Z \in \Ne{X}{\G} \setminus \{Y\}$, at least one of the following holds.
            \begin{enumerate}
                \item[] \textbf{Condition 1:\:}
                    $\exists \Wb \subseteq \Mb{X}{\G} \setminus \{Y,Z\}: \quad \sep{Y}{Z}{\Wb}{\G}$.
                \item[] \textbf{Condition 2:\:}
                    $\forall \Wb \subseteq \Mb{X}{\G} \setminus \{Y,Z\}: \quad \nsep{Y}{Z}{\Wb \cup \{X\}}{\G}$.
            \end{enumerate}
        \end{theorem}
        \begin{remark}
            If $Y \in \Ne{X}{\G}$, we can skip Condition 1 and only check Condition 2.
        \end{remark}
        Below, we present a proposition that we use in Section \ref{subsec: implementation L-MARVEL} to avoid performing duplicate CI tests in the implementation of L-MARVEL.

        \begin{proposition}\label{prp: L-MARVEL duplicate}
            Suppose $\G$ is a MAG with the set of vertices $\Vb$, $X \in \Vb$, $Y \in \Mb{X}{\G}$, and $Z \in \Ne{X}{\G}\setminus \{Y\}$.
            If at least one of the two conditions of Theorem \ref{thm: test removability} holds for $X,Y,Z$, then the graphical characterization for MAGs in Theorem \ref{thm: graph-rep MAG} also holds for $X,Y,Z$.
        \end{proposition}
    
    \subsection{RSL: Recursive Structure Learning}
        Another recursive algorithm for causal discovery is RSL, which aims to reduce the computational complexity of causal discovery under structural assumptions.
        RSL requires causal sufficiency and provides algorithms under two types of structural side information: (I) an upper bound on the clique number of the graph is known, or (II) the graph is diamond-free.
        The causal discovery algorithms provided under these assumptions are RSL$_{\omega}$ and RSL$_D$, respectively.
        Under the corresponding assumptions, both of these methods achieve polynomial-time complexity.
        
        \subsubsection{RSL\texorpdfstring{$_{\omega}$}{}}
            RSL$_{\omega}$ assumes that an upper bound $m$ on the clique number of the causal graph is known, i.e., $\omega(\G) \leq m$.
            \begin{remark} \label{remark: RSL_w erdos-renyi}
                For a random graph $\G$ generated from Erdos-Renyi model $G(n,p)$, $\omega(\G)\leq m$ with high probability when $p n^{2/m} \rightarrow 0$ as $n \rightarrow \infty$.
            \end{remark}
            RSL$_{\omega}$ provides the following result for verifying the removability of a variable under the assumption $\omega(\G) \leq m$.
            \begin{theorem} \label{thm: FindRemovable clique}
                Suppose $\G$ is a DAG such that $\omega(\G)\leq m$.
                Vertex $X$ is removable in $\G$ if for any $\Sb \subseteq \Mb{X}{\G}$ with $|\mathbf{S}| \leq m-2$, the following conditions hold.
                \begin{enumerate}
                    \item[] \textbf{Condition 1:}
                        $\nsep{Y}{Z}{\big(\Mb{X}{\G} \cup \{X\}\big) \setminus\big(\{Y,Z\} \cup \Sb \big)}{\G}, \quad \forall Y,Z \in \Mb{X}{\G} \setminus \Sb$.
                    \item[] \textbf{Condition 2:}
                        $\nsep{X}{Y}{\Mb{X}{\G} \setminus (\{Y\} \cup \Sb)}{\G}, \quad \forall Y \in \Mb{X}{\G} \setminus \Sb$.
                \end{enumerate}
                Also, the set of vertices that satisfy these conditions is nonempty.
            \end{theorem}
            To identify the neighbors of a removable variable detected using Theorem \ref{thm: FindRemovable clique}, RSL$_{\omega}$ provides the following proposition that distinguishes co-parents from neighbors in the Markov boundary.
            
            \begin{proposition} \label{prp: FindNeighbors clique}
                Suppose $\G$ is a DAG such that $\omega(\G)\leq m$.
                Let $X$ be a vertex that satisfies the two conditions of Theorem \ref{thm: FindRemovable clique} and $Y\in \Mb{X}{\G}$. 
                Then, $Y\in \CP{X}{\G}$ if and only if
                \begin{equation*}
                    \exists \Sb \subseteq \Mb{X}{\G} \setminus\{Y\}:\quad |\Sb|=m-1\,,\ \ \sep{X}{Y}{\Mb{X}{\G}\setminus(\{Y\}\cup\Sb)}{\G}.
                \end{equation*}
                Moreover, set $\Sb$ is unique and $\Sb=\Ch{X}{\G}\cap\Ch{Y}{\G}$.
            \end{proposition}
            
        \subsubsection{RSL\texorpdfstring{$_D$}{}}
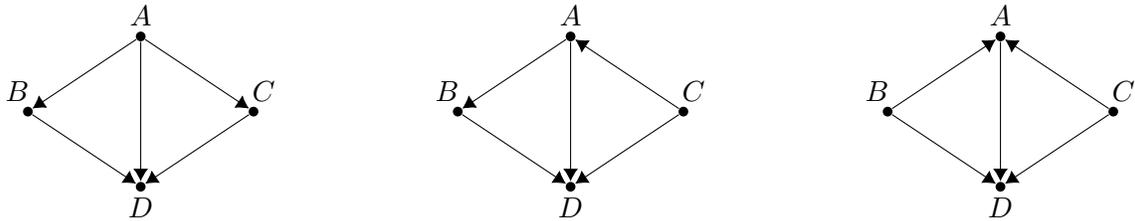
\begin{figure}[t!]
                \centering
                \tikzstyle{block} = [circle, inner sep=1.3pt, fill=black]
                \tikzstyle{input} = [coordinate]
                \tikzstyle{output} = [coordinate]
                \begin{subfigure}[b]{0.25\textwidth}
                    \centering
                    \begin{tikzpicture}
                        \tikzset{edge/.style = {->,> = latex',-{Latex[width=2mm]}}}
                        \node[block] (A) at  (0,2) {};
                        \node[] ()[above = -0.06cm of A]{$A$};
                        \node[block] (B) at  (-1.5,1) {};
                        \node[] ()[above left = -0.05cm and -0.2 cm of B]{$B$};
                        \node[block] (C) at  (1.5,1) {};
                        \node[] ()[above right = -0.05cm and -0.2 cm of C]{$C$};
                        \node[block] (D) at  (0,0) {};
                        \node[] ()[below = -0.06cm of D]{$D$};
                        \draw[edge] (A) to (B);
                        \draw[edge] (A) to (C);
                        \draw[edge] (A) to (D);
                        \draw[edge] (B) to (D);
                        \draw[edge] (C) to (D);
                    \end{tikzpicture}
                \end{subfigure} \hfill
                \begin{subfigure}[b]{0.25\textwidth}
                    \centering
                    \begin{tikzpicture}
                        \tikzset{edge/.style = {->,> = latex',-{Latex[width=2mm]}}}
                        \node[block] (A) at  (0,2) {};
                        \node[] ()[above = -0.06cm of A]{$A$};
                        \node[block] (B) at  (-1.5,1) {};
                        \node[] ()[above left = -0.05cm and -0.2 cm of B]{$B$};
                        \node[block] (C) at  (1.5,1) {};
                        \node[] ()[above right = -0.05cm and -0.2 cm of C]{$C$};
                        \node[block] (D) at  (0,0) {};
                        \node[] ()[below = -0.06cm of D]{$D$};
                        \draw[edge] (A) to (B);
                        \draw[edge] (C) to (A);
                        \draw[edge] (A) to (D);
                        \draw[edge] (B) to (D);
                        \draw[edge] (C) to (D);
                    \end{tikzpicture}
                \end{subfigure} \hfill         
                \begin{subfigure}[b]{0.25\textwidth}
                \centering
                    \begin{tikzpicture}
                        \tikzset{edge/.style = {->,> = latex',-{Latex[width=2mm]}}}
                        \node[block] (A) at  (0,2) {};
                        \node[] ()[above = -0.06cm of A]{$A$};
                        \node[block] (B) at  (-1.5,1) {};
                        \node[] ()[above left = -0.05cm and -0.2 cm of B]{$B$};
                        \node[block] (C) at  (1.5,1) {};
                        \node[] ()[above right = -0.05cm and -0.2 cm of C]{$C$};
                        \node[block] (D) at  (0,0) {};
                        \node[] ()[below = -0.06cm of D]{$D$};
                        \draw[edge] (B) to (A);
                        \draw[edge] (C) to (A);
                        \draw[edge] (A) to (D);
                        \draw[edge] (B) to (D);
                        \draw[edge] (C) to (D);
                    \end{tikzpicture}
                \end{subfigure}
                \caption{Diamond graphs.}
                \label{fig: diamond}
            \end{figure}

            RSL$_D$ assumes that the causal DAG is diamond-free, i.e., it contains no diamond as an induced subgraph.
            Diamond is one of the three types of DAGs shown in Figure \ref{fig: diamond}.
            
            \begin{remark} \label{remark: RSL_D erdos-renyi}
                A random graph $\G$ generated from Erdos-Renyi model $G(n,p)$ is diamond-free with high probability when $pn^{0.8}\!\rightarrow\!0$.
            \end{remark}
            The following theorem provides an efficient method for checking the removability of a variable when the causal DAG is diamond-free.
            
            \begin{theorem} \label{thm: FindRemovable RSL_D}
                In a diamond-free DAG $\G$, a vertex $X$ is removable if and only if
                \begin{equation*}
                    \nsep{Y}{Z}{(\Mb{X}{\G} \cup \{X\})\setminus \{Y,Z\}}{\G}, \quad \forall Y,Z \in \Mb{X}{\G}.
                \end{equation*}
            \end{theorem}
            Analogous to the case with the bounded clique number, the following proposition can be used to learn the neighbors of a removable variable in a diamond-free DAG.

            \begin{proposition} \label{prp: FindNeighbors RSL_D}
                In a diamond-free DAG $\G$, let $X$ be a removable variable and $Y\in \Mb{X}{\G}$.
                In this case, $Y \in \CP{X}{\G}$ if and only if 
                \begin{equation}\label{eq:v-rsld}
                    \exists Z\in \Mb{X}{\G}\setminus \{Y\}:\quad  \sep{X}{Y}{\Mb{X}{\G} \setminus \{Y,Z\}}{\G}.
                \end{equation}
                Moreover, such a variable $Z$ is unique and $\{Z\}=\Ch{X}{\G}\cap\Ch{Y}{\G}$.
            \end{proposition}

    \subsection{Removable-Order Learning: ROL} \label{sec: ROL}
        ROL is an ordering-based method that leverages the notion of removability for causal discovery and does not require the assumption of causal sufficiency.
        As discussed in Section \ref{subsec: removable orders}, ordering-based methods in the literature prior to this approach recover a graph through learning a causal order (c-order) of DAGs, which is a topological order of variables (Definition \ref{def: c-order}).
        ROL introduces and uses a novel order called removable order (r-order), which we defined for MAGs in Definition \ref{def: r-order}.
        
        Note that in our general framework given in Algorithm \ref{alg: Recursive framework}, $(X_1,\dots, X_n)$ forms an r-order.
        While the recursive methods that we discussed in the previous subsections seek to identify a removable variable in each iteration, ROL aims to learn the whole order at once.
        To this end, ROL first proposes a recursive algorithm that learns an undirected graph, whose pseudocode is given in Algorithm \ref{alg: Gpi}.

        \begin{algorithm}[ht]
            \caption{Learning $\Gpi$.}
            \label{alg: Gpi}
            \begin{algorithmic}[1]
                \STATE \textbf{Function LearnGPi} ($\pi,\, \D(\Vb)$)
                \STATE $\Vb_1 \gets \Vb$,  $\Eb^{\pi}\gets \varnothing$
                \FOR{$t=1$ to $n-1$}
                    \STATE $X_t \gets \pi(t)$
                    \STATE $\Ne{X_t}{\Vb_t} \gets \textbf{FindNeighbors}(X_t, \D(\Vb_t))$
                    \STATE Add undirected edges between $X_t$ and the variables in $\Ne{X_t}{\Vb_t}$ to $\Eb^{\pi}$.
                    \STATE $\Vb_{t+1} \gets \Vb_t \setminus \{X_t\}$
                \ENDFOR
                \STATE \textbf{Return} $\Gpi = (\Vb, \Eb^{\pi})$
            \end{algorithmic}
        \end{algorithm}
        
        Algorithm \ref{alg: Gpi} removes variables according to an arbitrary given order $\pi$ and learns an undirected graph $\G^{\pi}$ recursively.
        Using this algorithm, ROL defines the cost of an order $\pi$ to be the number of edges in $\Gpi$, denoted by $|\Eb^{\pi}|$.
        It then defines an optimization problem that seeks to find an order with the minimum cost.
        
        \begin{theorem}[Consistency] \label{thm: opt prob}
            Any solution of the optimization problem
            \begin{equation}\label{eq: opt problem}
                \argmin_{\pi} |\Eb^{\pi}|,
            \end{equation}
            is a member of $\Pi^r(\G)$.
            Conversely every member of $\Pi^r(\G)$ is a solution of \eqref{eq: opt problem}.
        \end{theorem}
        Theorem \ref{thm: opt prob} shows that finding an r-order is equivalent to solving \eqref{eq: opt problem}.
        To solve this optimization problem, ROL proposes three algorithms.
        \begin{itemize}
            \item ROL$_{\text{HC}}$, a Hill-climbing-based heuristic algorithm that is scalable to large graphs.
            \item ROL$_{\text{VI}}$, an exact reinforcement learning (RL)-based algorithm that has theoretical guarantees but is not scalable to large graphs.
            \item ROL$_{\text{PG}}$, an approximate RL-based algorithm that exploits stochastic policy gradient.
        \end{itemize}
        In Section \ref{subsec: implementation details - ROL}, we will delve into the details of these methods.
        Herein, we discuss how ROL formulates the optimization problem in Theorem \ref{thm: opt prob} as an RL problem.
        
        ROL interprets the process of recovering $\mathcal{G}^\pi$ from a given order $\pi$ as a Markov decision process (MDP), where the index $t$ denotes time, the action space is the set of variables $\Vb$, and the state space is the set of all subsets of $\Vb$.
        More precisely, let $s_t$ and $a_t$ denote the state and the action of the MDP at iteration $t$, respectively.
        Then, $s_t$ is the remaining variables at time $t$ (i.e., $s_t=\Vb_t$) and action $a_t$ is the variable that will be removed from $\Vb_t$ in that iteration (i.e., $a_t=X_t$).
        Consequently, the state transition due to action $a_t$ is $s_{t+1}=\Vb_t \setminus \{a_t\}$.
        The immediate reward of selecting action $a_t$ at state $s_t$ will be the negative of the instantaneous cost, naturally defined as the number of discovered neighbors for $a_t$ by \textbf{FindNeighbors} in line 5 of Algorithm \ref{alg: Gpi}.
        Formally, the reward of picking action $a_t$ when in state $s_t$ is thus given by
        \begin{align*}
            r\left(s_t, a_t\right)=| \textbf{FindNeighbors}\left(a_t, \operatorname{Data}\left(s_t\right)\right)| = - \left|\Ne{a_t}{s_t}\right|.
        \end{align*}
        In Sections \ref{sec: implementation - ROL_VI} and \ref{sec: implementation - ROL_PG}, we discuss two of our RL-based approaches with the above formulation.

\section{Implementation Details} \label{sec: implementation details}
    In Section \ref{sec: RCD methods}, we explored techniques for identifying removable variables that can be used in various recursive causal discovery algorithms.
    However, certain details were left out.
    In this section, we will discuss implementation details and provide pseudocode for these methods.
    Moreover, in Section \ref{sec: MEC}, we discuss how our methods can be augmented to identify the MEC of the underlying causal graph.
    \subsection{MARVEL}   
        Algorithm \ref{alg: MARVEL} provides a pseudocode for MARVEL, which is compatible with the generic frameworks of Algorithms \ref{alg: Recursive framework} and \ref{alg: FindRemovable}.
        Algorithm \ref{alg: MARVEL functions} presents the main functions that MARVEL uses to learn neighbors and v-structures, as well as to verify the removability of a variable.
        As elaborated in Section \ref{sec: duplicate CI tests MARVEL}, MARVEL incorporates three data structures - $\F$, $\FI$, and $\FII$.
        These structures are designed to avoid performing redundant CI tests and improve overall computational efficiency.
        
        \begin{algorithm}[t!]
            \caption{MARVEL}
            \label{alg: MARVEL}
            \begin{algorithmic}[1]
                \STATE {\bfseries Input:} $\D(\Vb)$
                \STATE Initialize undirected graph $\hat{\G} = (\Vb,\,\Eb=\varnothing)$
                \STATE $\Vb_1 \gets \Vb$
                \STATE $\forall X \in \Vb\!:\: \F(X) \gets$ \false
                \STATE $\forall X,Y,Z \in \Vb\!:\: \FI(X,Y,Z),\,\FII(X,Y,Z) \gets$ \false
                \STATE $\MB{\Vb_1} \gets$ \textbf{ComputeMb}($\D(\Vb)$)
    		\FOR{$i$ from $1$ to $n-1$}
                    \STATE $\Tilde{\Vb}_i \gets \{X \in \Vb_i \vert  \F(X_j) = \false\} $
                    \STATE $r \gets |\Tilde{\Vb}_i|$
    		      \STATE $(X_1, \dots, X_r) \gets$ Sort $\Tilde{\Vb}_i$ such that $|\Mb{X_1}{\Vb_i}| \leq |\Mb{X_2}{\Vb_i}|\leq \dots \leq |\Mb{X_r}{\Vb_i}|$
    		    \FOR{$j$ from $1$ to $r$}\label{line: FindRemovable MARVEL}
                        \IF{First time applying the \textbf{FindNeighbors} function to $X_j$}
                            \STATE $\Ne{X_j}{\Vb_i},\,\CP{X_j}{\Vb_i},\,\sepset(X_j) \gets$ \textbf{FindNeighbors($X_j,\,\Mb{X_j}{\Vb_i}$)}
                            \STATE Add undirected edges between $X_j$ and $\Ne{X_j}{\Vb_i}$ in $\hat{\G}$ if not present
                        \ELSE
                            \STATE $\Ne{X_j}{\Vb_i} \gets \Ne{X_j}{\hat{\G}[\Vb_i]}$
                            \STATE $\CP{X_j}{\Vb_i} \gets \Mb{X_j}{\Vb_i} \setminus \Ne{X_j}{\Vb_i}$
                        \ENDIF
                        \IF{\textbf{Condition1}($X_j,\,\Ne{X_j}{\Vb_i},\,\Mb{X_j}{\Vb_i}$) is \true}
                            \IF{First time applying the \textbf{FindVS} function to $X_j$}
                                \STATE $\VS{X_j}{} \gets $ \textbf{FindVS}($X_j,\,\Ne{X_j}{\Vb_i},\,\CP{X_j}{\Vb_i},\,\Mb{X_j}{\Vb_i},\,\sepset(X_j)$)
                            \ELSE
                                \STATE $\VS{X_j}{} \gets \{(X_j \to Z \gets Y) \in \VS{X_j}{} \vert Z,Y \in \Vb_i \}$
                            \ENDIF
                            \IF{\textbf{Condition2}(($X_j,\,\Ne{X_j}{\Vb_i},\,\CP{X_j}{\Vb_i},\,\Mb{X_j}{\Vb_i},\,\VS{X_j}{}$)) is \true}
                                \STATE $\Vb_{i+1} \gets \Vb_i \setminus \{X_j\}$
                                \STATE $\MB{\Vb_{i+1}} \gets$ \textbf{UpdateMb}($X_j,\,\Ne{X_j}{\Vb_i},\,\MB{\Vb_i}$)
                                \STATE \textbf{Break} the for loop of line \ref{line: FindRemovable MARVEL}
                            \ELSE
                                \STATE $\F \gets \true$
                            \ENDIF
                        \ELSE
                            \STATE $\F \gets \true$
                        \ENDIF                        
                    \ENDFOR
                \ENDFOR
                \RETURN $\hat{\G}$
            \end{algorithmic}
        \end{algorithm}

        \begin{algorithm}[t!]
            \caption{MARVEL functions}
            \label{alg: MARVEL functions}
            \begin{algorithmic}[1]
                \STATE {\bfseries Function FindNeighbors}($X,\,\Mb{X}{\Wb}$) \hfill \% Lemma \ref{lem: neighbor}
                \STATE $\Ne{X}{\Wb},\, \CP{X}{\Wb},\, \sepset(X) \gets \varnothing$
                \FOR{$Y \in \Mb{X}{\Wb}$}
                    \IF{$\exists \Sb_{X,Y} \subsetneq \Mb{X}{\Wb} \setminus \{Y\}: \quad \sep{X}{Y}{\Sb_{X,Y}}{\D(\Wb)}$}
                        \STATE Add $Y$ to $\CP{X}{\Wb}$
                        \STATE Add $\Sb_{X,Y}$ to $\sepset(X)$
                    \ELSE
                        \STATE Add $Y$ to $\Ne{X}{\Wb}$
                    \ENDIF
                \ENDFOR
                \STATE \textbf{return} $\Ne{X}{\Wb},\,\CP{X}{\Wb},\,\sepset(X)$
            \end{algorithmic}
            \hrulefill
            \begin{algorithmic}[1]
                \STATE {\bfseries Function FindVS}($X,\,\Ne{X}{\Wb},\,\CP{X}{\Wb},\,\Mb{X}{\Wb},\,\sepset(X)$) \hfill \% Lemma \ref{lem: v-structures}
                \STATE $\VS{X}{} \gets \varnothing$
                \FOR{$Y \in \CP{X}{\Wb}$ and $Z \in \Ne{X}{\Wb}$}
                    \STATE $\Sb_{X,Y} \gets$ The separating set in $\sepset$ corresponding to $Y$
                    \IF{$Z \notin \Sb_{XY}$ and $\forall \Sb \subseteq \Mb{X}{\Wb} \cup\{X\} \setminus \{Y,Z\}: \quad \nsep{Y}{Z}{\Sb}{\D(\Wb)}$}
                        \STATE Add $(X \to Z \gets Y)$ to $\VS{X}{}$
                    \ENDIF
                \ENDFOR
                \STATE \textbf{return} $\VS{X}{}$
            \end{algorithmic}
            \hrulefill
            \begin{algorithmic}[1]
                \STATE {\bfseries Function Condition1}($X,\,\Ne{X}{\Wb},\,\Mb{X}{\Wb}$) \hfill \% Lemma \ref{lem: condition1}
                \FOR{$Y, Z \in \Ne{X}{\Wb}$ such that $\FI(X,Y,Z)$ is \false}
                    \FOR{$\Sb \subseteq \Mb{X}{\Wb} \setminus \{Y, Z\}$}
                        \IF{$\sep{Y}{Z}{\Sb \cup \{X\}}{\D(\Wb)}$}
                            \STATE \textbf{return} \false
                        \ENDIF
                    \ENDFOR
                    \STATE $\FI(X,Y,Z) \gets \true$
                \ENDFOR
                \STATE \textbf{return} \true              
            \end{algorithmic}
            \hrulefill
            \begin{algorithmic}[1]
                \STATE {\bfseries Function Condition2}($X,\,\Ne{X}{\Wb},\,\CP{X}{\Wb},\,\Mb{X}{\Wb},\,\VS{X}{}$) \hfill \% Lemma \ref{lem: condition2}
                \FOR{$Y \in \CP{X}{\Wb}$ and $Z \in \Ne{X}{\Wb}$ such that $\FII(X,Y,Z)$ is \false}
                    \STATE $\boldsymbol{\Gamma}\gets\{V\neq Z:(X \to V\gets Y) \in \VS{X}{}\}$
                    \FOR{$\Sb \subseteq \Mb{X}{\Wb} \setminus\{ Y, Z\}$ s.t. $\Sb\cap\boldsymbol{\Gamma}\neq\emptyset$}
                        \IF{$\sep{Y}{Z}{\Sb \cup\{X\}}{\D(\Wb)}$}
                            \STATE \textbf{return} \false
                        \ENDIF
                    \ENDFOR
                    \STATE $\FII(X,Y,Z) \gets \true$
                \ENDFOR
                \STATE \textbf{return} \true              
            \end{algorithmic}
        \end{algorithm}

        Algorithm \ref{alg: MARVEL} initializes the required variables in lines 2--5.
        It then calls the \textbf{ComputeMb} function to initially compute the Markov boundaries.
        We will discuss this step in Section \ref{sec: ComputeMb function}.
        In the for loop of lines 11--30, the removability of variable $X_j$ is checked, and its neighbors among the remaining variables are learned.
        If the neighbors of $X_j$ have not been learned in the previous iterations, the algorithm calls the \textbf{FindNeighbors} function in Algorithm \ref{alg: MARVEL functions} to learn its neighbors.
        This function uses Lemma \ref{lem: neighbor} to learn the following.
        \begin{equation*}
            \Ne{X_j}{\Vb_i}, \quad 
            \CP{X_j}{\Vb_i}, \quad
            \{\Sb_{X_jY}: Y\in \CP{X_j}{\Vb_i}, \sep{X_j}{Y}{\Sb_{X_jY}}{P_{\Vb_i}}\}.
        \end{equation*}
        If the neighbors of $X_j$ have been learned in previous iterations, the algorithm uses $\hat{\G}$ to recover this information.
        Then, upon calling function \textbf{Condition1}, we verify the first condition of Theorem \ref{thm: graph-rep DAG}, which is designed in accordance with Lemma \ref{lem: condition1}.
        If this condition is satisfied, Lemma \ref{lem: v-structures} is applied (function \textbf{FindVS}) to learn $\VS{X_j}{\Vb_i}$.
        Note that if $\VS{X}{\Vb_i}$ is learned using Lemma \ref{lem: v-structures} in an iteration, we can save it and delete a v-structure from it when one of the three variables of the v-structure is removed.
        Finally, the second condition of Theorem \ref{thm: graph-rep DAG} is verified using Lemma \ref{lem: condition2} (function \textbf{Condition2}).
        If this condition is also satisfied, the algorithm concludes that $X_j$ is removable and proceeds to remove it from the variables.
        For the next iteration, we update the Markov boundaries by calling the \textbf{UpdateMb} function in Algorithm \ref{alg: update Mb}, which we discuss in Section \ref{sec: UpdateMb}.

        \subsubsection{Avoiding Duplicate CI Tests in MARVEL} \label{sec: duplicate CI tests MARVEL}
            Suppose that MARVEL verifies the removability of a variable $X$ and determines that it is not removable.
            As a result, the algorithm will need to verify again the removability of $X$ in some of the following iterations, potentially leading to redundant CI tests.
            To address this, we propose a method to eliminate such redundancies by leveraging information from previous iterations.

            Suppose that during iteration $i$, we invoke \textbf{Condition1} from Algorithm \ref{alg: MARVEL functions} for a variable $X$, where $\Wb = \Vb_i$.
            If two variables $Y, Z\in \Ne{X}{\Vb_i}$ do not have a separating set in $\Mb{X}{\Vb_i} \setminus \{Y, Z\}$, then they will not have a separating set in $\Mb{X}{\Vb_{i'}} \setminus \{Y, Z\}$ for any $i'>i$.
            Accordingly, to prevent redundant CI tests, \FI is employed in function \textbf{Condition1} to save this information and avoid performing duplicate CI tests.
            A similar approach is adopted in \textbf{Condition2} using \FII.

            To further enhance the implementation, \F is integrated into Algorithm \ref{alg: MARVEL}.
            This is based on the understanding that a variable's removability hinges on its Markov boundary.
            If a variable $X$ is found non-removable in one iteration, it remains so as long as its Mb is unchanged, thereby obviating the need to recheck within the for loop of lines 11--30.
            Initially, every variable in $\Vb$ has \F set to \false. 
            Should either function \textbf{Condition1} or \textbf{Condition2} return \false, we switch $\F(X)$ to \true.
            $\F(X)$ stays \true unless there is a change in its Mb. 
            As detailed in Section \ref{sec: UpdateMb}, updating \F is crucial and is done in the \textbf{UpdateMb} function.

    \subsection{L-MARVEL} \label{subsec: implementation L-MARVEL}
        Similar to MARVEL, we provide a pseudocode for L-MARVEL in Algorithm \ref{alg: L-MARVEL}.
        
        \begin{algorithm}[p]
            \caption{L-MARVEL}
            \label{alg: L-MARVEL}
            \begin{algorithmic}[1]
                \STATE {\bfseries Input:} $\D(\Vb)$
                \STATE Initialize undirected graph $\hat{\G} = (\Vb,\,\Eb=\varnothing)$
                \STATE $\forall X \in \Vb\!:\: \F(X) \gets$ \false
                \STATE $\forall X,Y,Z \in \Vb\!:\: \FF(X,Y,Z) \gets$ \false
                \STATE $\Vb_1 \gets \Vb$
                \STATE $\MB{\Vb_1} \gets$ \textbf{ComputeMb}($\D(\Vb)$)
    		    \FOR{$i$ from $1$ to $n-1$}
                        \STATE $\Tilde{\Vb}_i \gets \{X \in \Vb_i \vert  \F(X_j) = \false\} $
                        \STATE $r \gets |\Tilde{\Vb}_i|$
    		          \STATE $(X_1, \dots, X_r) \gets$ Sort $\Tilde{\Vb}_i$ such that $|\Mb{X_1}{\Vb_i}| \leq |\Mb{X_2}{\Vb_i}|\leq \dots \leq |\Mb{X_r}{\Vb_i}|$
    		        \FOR{$j$ from $1$ to $r$} \label{line: FindRemovable L-MARVEL}
                            \IF{First time learning the neighbors of $X_j$}
                                \STATE $\Ne{X_j}{\Vb_i} \gets$ \textbf{FindNeighbors}($X_j,\,\Mb{X_j}{\Vb_i}$)
                                \STATE Add undirected edges between $X_j$ and $\Ne{X_j}{\Vb_i}$ in $\hat{\G}$ if not present
                            \ELSE
                                \STATE $\Ne{X_j}{\Vb_i} \gets \Ne{X_j}{\hat{\G}[\Vb_i]}$ 
                            \ENDIF
                            \IF{\textbf{IsRemovable}($X_j,\,\Ne{X_j}{\Vb_i},\,\Mb{X_j}{\Vb_i})$) is \true}
                                \STATE $\Vb_{i+1} \gets \Vb_i \setminus \{X_j\}$
                                \STATE $\MB{\Vb_{i+1}} \gets$ \textbf{UpdateMb}($X_j,\,\Ne{X_j}{\Vb_i},\,\MB{\Vb_i}$)
                                \STATE \textbf{Break} the for loop of line \ref{line: FindRemovable L-MARVEL}
                            \ELSE
                                \STATE $\F(X_j) \gets $ \true
                            \ENDIF
    		        \ENDFOR
    		    \ENDFOR
    		    \RETURN $\hat{\G}$
            \end{algorithmic}
            \hrulefill
            \begin{algorithmic}[1]
                \STATE {\bfseries Function FindNeighbors}($X,\,\Mb{X}{\Wb}$) \hfill \% Lemma \ref{lem: neighbor}
                \STATE $\Ne{X}{\Wb} \gets \Mb{X}{\Wb}$
                \FOR{$Y \in \Mb{X}{\Wb}$}
                    \IF{$\exists \Sb \subsetneq \Mb{X}{\Wb} \setminus \{Y\}: \quad \sep{X}{Y}{\Sb}{\D(\Wb)}$}
                        \STATE Remove $Y$ from $\Ne{X}{\Wb}$
                    \ENDIF
                \ENDFOR
                \STATE \textbf{return} $\Ne{X}{\Wb}$
            \end{algorithmic}
            \hrulefill
            \begin{algorithmic}[1]
                \STATE {\bfseries Function IsRemovable}($X,\,\Ne{X}{\Wb},\, \Mb{X}{\Wb}$) \hfill \% Theorem \ref{thm: test removability}
                \FOR{$Y\in \Mb{X}{\Wb}$ and $Z\in \Ne{X}{\Wb}$ such that $\FF(X,Y,Z)$ is \false}
                    \IF{$\forall \Sb \subseteq \Mb{X}{\Wb} \setminus \{Y,Z\}: \quad \nsep{Y}{Z}{\Sb}{\D(\Wb)}$}
                        \IF{$\exists \Sb \subseteq \Mb{X}{\Wb} \setminus \{Y,Z\}: \quad \sep{Y}{Z}{\Sb \cup \{X\}}{\D(\Wb)}$}
                            \STATE \textbf{Return} \false
                        \ENDIF
                    \ENDIF
                    \STATE $\FF(X,Y,Z) \gets \true$
                \ENDFOR
                \STATE \textbf{Return} \true
            \end{algorithmic}
        \end{algorithm}

        In the for loop of lines 11--22, the algorithm learns the neighbors of $X_j$ and verifies its removability.
        If the algorithm is learning the neighbors of $X_j$ for the first time, it uses Lemma \ref{lem: neighbor} (function \textbf{FindNeighbors}) to learn $\Ne{X_j}{\Vb_i}$.   
        Otherwise, this information has already been stored in $\hat{\G}$, and the algorithm executes line $13$ to recover the neighbors of $X_j$ among the remaining variables.
        To verify the removability of $X_j$, the \textbf{IsRemovable} function is invoked.
        This function checks the two conditions of Theorem \ref{thm: test removability}.
        If both conditions are not met, the function determines that $X_j$ is not removable.

        Algorithm \ref{alg: L-MARVEL} uses \F and \FF to avoid performing duplicate CI tests.
        Similar to MARVEL, if a variable is found non-removable in one iteration, its \F value is set to \true, and it remains so as long as the Markov boundary of the variable remains unchanged.
        Additionally, Proposition \ref{prp: L-MARVEL duplicate} implies that if the condition of Theorem \ref{thm: test removability} is met for $Y$ and $Z$ during an iteration in the \textbf{IsRemovable} function, the graphical characterization of Theorem \ref{thm: graph-rep MAG} holds for $X,Y,Z$.
        Therefore, to check the removability of $X$, it is not necessary to check the two conditions of Theorem \ref{thm: test removability}.
        Accordingly, \FF is employed in the \textbf{IsRemovable} function to avoid performing redundant CI tests.
    
    \subsection{RSL}
        In this part, we discuss the implementation details of RSL$_{\omega}$ and RSL$_{D}$.
        \subsubsection{RSL\texorpdfstring{$_{\omega}$}{}}
            Algorithm \ref{alg: RSL_w} presents the pseudocode for RSL$_{\omega}$.
            This algorithm takes $m$ as an input, which is an upper bound on the clique number of the true underlying graph.
            One of the differences between this algorithm and MARVEL and L-MARVEL is the sequence of operations.
            This algorithm first checks the removability of a variable using Theorem \ref{thm: FindRemovable clique} through the function \textbf{IsRemovable}.
            Subsequently, it finds the neighbors of the variable by employing Proposition \ref{prp: FindNeighbors clique} in the function \textbf{FindNeighbors}.
            Furthermore, Algorithm \ref{alg: RSL_D} only uses one data structure, \F, to efficiently prevent redundant CI tests.

            \begin{figure}[p]
                \begin{minipage}{\textwidth}
                    \begin{algorithm}[H]
                        \caption{RSL$_{\omega}$}
                        \label{alg: RSL_w}
                        \begin{algorithmic}[1]
                            \STATE {\bfseries Input:} $\D(\Vb)$, $m$
                            \STATE Initialize undirected graph $\hat{\G} = (\Vb,\,\Eb=\varnothing)$
                            \STATE $\forall X \in \Vb\!:\: \F(X) \gets$ \false
                            \STATE $\Vb_1 \gets \Vb$
                            \STATE $\MB{\Vb_1} \gets$ \textbf{ComputeMb}($\D(\Vb)$)
                            \FOR{$i$ from $1$ to $n-1$}
                                \STATE $\Tilde{\Vb}_i \gets \{X \in \Vb_i \vert  \F(X_j) = \false\} $
                                \STATE $r \gets |\Tilde{\Vb}_i|$
                              \STATE $(X_1, \dots, X_r) \gets$ Sort $\Tilde{\Vb}_i$ such that $|\Mb{X_1}{\Vb_i}| \leq |\Mb{X_2}{\Vb_i}|\leq \dots \leq |\Mb{X_r}{\Vb_i}|$
                            \FOR{$j$ from $1$ to $r$} \label{line: FindRemovable RSL_W}
                                    \IF{\textbf{IsRemovable}($X_j,\,\Mb{X_j}{\Vb_i}),\,m$) is \true}
                                        \STATE $\Ne{X_j}{\Vb_i} \gets$ \textbf{FindNeighbors}($X_j,\,\Mb{X_j}{\Vb_i},\,m$)
                                        \STATE Add undirected edges between $X_j$ and $\Ne{X_j}{\Vb_i}$ in $\hat{\G}$ if not present
                                        \STATE $\Vb_{i+1} \gets \Vb_i \setminus \{X_j\}$
                                        \STATE $\MB{\Vb_{i+1}} \gets$ \textbf{UpdateMb}($X_j,\,\Ne{X_j}{\Vb_i},\,\MB{\Vb_i}$)
                                        \STATE \textbf{Break} the for loop of line \ref{line: FindRemovable RSL_W}
                                    \ELSE
                                        \STATE $\F(X_j) \gets $ \true
                                    \ENDIF
                                \ENDFOR
                            \ENDFOR
                        \RETURN $\hat{\G}$
                        \end{algorithmic}
                        \hrulefill
                        \begin{algorithmic}[1]
                            \STATE {\bfseries Function IsRemovable}($X,\,\Mb{X}{\Wb},\,m$) \hfill \% Theorem \ref{thm: FindRemovable clique}
                            \FOR{$\Sb \subseteq \Mb{X}{\Wb}$ with $|\Sb|\leq m-2$}
                                \IF{$\left(\exists Y,Z \in \Mb{X}{\Wb} \setminus \Sb: \quad \sep{Y}{Z}{\big(\Mb{X}{\Wb} \cup \{X\}\big) \setminus\big(\{Y,Z\} \cup \Sb \big)}{\D(\Wb)} \right)$ \\
                                or $\left( \exists Y \in \Mb{X}{\Wb} \setminus \Sb: \quad \sep{X}{Y}{\Mb{X}{\Wb} \setminus (\{Y\} \cup \Sb)}{\D(\Wb)} \right)$}
                                    \STATE \textbf{Return} \false
                                \ENDIF
                            \ENDFOR
                            \STATE \textbf{Return} \true
                        \end{algorithmic}
                        \hrulefill
                        \begin{algorithmic}[1]
                            \STATE {\bfseries Function FindNeighbors}($X,\,\Mb{X}{\Wb},\,m$) \hfill \% Proposition \ref{prp: FindNeighbors clique}
                            \STATE $\Ne{X}{\Wb} \gets \Mb{X}{\Wb}$
                            \FOR{$Y \in \Mb{X}{\Wb}$}
                                \IF{$\exists \Sb \subseteq \Mb{X}{\Wb} \setminus\{Y\}: |\Sb|=m-1, \sep{X}{Y}{\Mb{X}{\Wb}\setminus(\{Y\}\cup\Sb)}{\D(\Wb)}$}
                                    \STATE Remove $Y$ from $\Ne{X}{\Wb}$
                                \ENDIF
                            \ENDFOR
                            \STATE \textbf{return} $\Ne{X}{\Wb}$
                        \end{algorithmic}
                    \end{algorithm}
                \end{minipage}
                \begin{minipage}{\textwidth}
                    \begin{algorithm}[H]
                        \caption{RSL$_{\omega}$ Without Side Information.}
                        \label{alg: without side information}
                        \begin{algorithmic}[1]
                            \STATE {\bfseries Input:} $\D(\Vb)$
                            \FOR{$m$ from $1$ to $n$}
                                \STATE $\hat{\G} \gets \textbf{RSL}_{\omega}(\D(\Vb), m)$
                                \IF{\textbf{RSL}$_{\omega}$ terminates and $\omega(\hat{\G})\leq m$}
                                    \RETURN $\hat{\G}$
                                \ENDIF
                            \ENDFOR
                        \end{algorithmic}
                    \end{algorithm}
                \end{minipage}
            \end{figure}
        
            As mentioned above, Algorithm \ref{alg: RSL_w} takes $m$, an upper bound on the clique number of the causal graph.
            But what happens if it is not a valid upper bound, i.e., $m < \omega(\G)$?
            In this scenario, two outcomes are possible: either Algorithm \ref{alg: RSL_w} is unable to identify any removable variables at an iteration and halts, or RSL$_\omega$ terminates and returns a graph.
            \begin{proposition}[RSL$_{\omega}$ is verifiable] \label{prp: verifiable}
                If Algorithm \ref{alg: RSL_w} terminates with an input $m>0$, then the clique number of the learned skeleton is greater than or equal to the clique number of the true causal graph.
            \end{proposition}
            Proposition \ref{prp: verifiable} implies that, upon termination of the algorithm, if the clique number of the learned graph is at most $m$, then $m$ is a valid upper bound on the clique number, ensuring the correctness of the output.
            Otherwise, it indicates that the true clique number is greater than $m$.
            Accordingly, we present Algorithm \ref{alg: without side information} that can learn the skeleton of $\G$ without any prior knowledge about $\omega(\G)$.

            \begin{algorithm}[t!]
                \caption{RSL$_D$}
                \label{alg: RSL_D}
                \begin{algorithmic}[1]
                    \STATE {\bfseries Input:} $\D(\Vb)$
                    \STATE Initialize undirected graph $\hat{\G} = (\Vb,\,\Eb=\varnothing)$
                    \STATE $\forall X \in \Vb\!:\: \F(X) \gets$ \false
                    \STATE $\Vb_1 \gets \Vb$
                    \STATE $\MB{\Vb_1} \gets$ \textbf{ComputeMb}($\D(\Vb)$)
                    \FOR{$i$ from $1$ to $n-1$}
                        \STATE $\Tilde{\Vb}_i \gets \{X \in \Vb_i \vert  \F(X_j) = \false\} $
                        \STATE $r \gets |\Tilde{\Vb}_i|$
    		          \STATE $(X_1, \dots, X_r) \gets$ Sort $\Tilde{\Vb}_i$ such that $|\Mb{X_1}{\Vb_i}| \leq |\Mb{X_2}{\Vb_i}|\leq \dots \leq |\Mb{X_r}{\Vb_i}|$
    		        \FOR{$j$ from $1$ to $r$} \label{line: FindRemovable RSL_D}
                            \IF{\textbf{IsRemovable}($X_j,\,\Mb{X_j}{\Vb_i}$) is \true}
                                \STATE $\Ne{X_j}{\Vb_i} \gets$ \textbf{FindNeighbors}($X_j,\,\Mb{X_j}{\Vb_i}$)
                                \STATE Add undirected edges between $X_j$ and $\Ne{X_j}{\Vb_i}$ in $\hat{\G}$ if not present
                                \STATE $\Vb_{i+1} \gets \Vb_i \setminus \{X_j\}$
                                \STATE $\MB{\Vb_{i+1}} \gets$ \textbf{UpdateMb}($X_j,\,\Ne{X_j}{\Vb_i},\,\MB{\Vb_i}$)
                                \STATE \textbf{Break} the for loop of line \ref{line: FindRemovable RSL_D}
                            \ELSE
                                \STATE $\F(X_j) \gets $ \true
                            \ENDIF
                        \ENDFOR
                    \ENDFOR
            	\RETURN $\hat{\G}$
                \end{algorithmic}
                \hrulefill
                \begin{algorithmic}[1]
                    \STATE {\bfseries Function IsRemovable}($X,\,\Mb{X}{\Wb}$) \hfill \% Theorem \ref{thm: FindRemovable RSL_D}
                    \FOR{$Y,Z \in \Mb{X}{\Wb}$}
                        \IF{$\sep{Y}{Z}{(\Mb{X}{\Wb} \cup \{X\})\setminus \{Y,Z\}}{\D(\Wb)}$}
                            \STATE \textbf{Return} \false
                        \ENDIF
                    \ENDFOR
                    \STATE \textbf{Return} \true
                \end{algorithmic}
                \hrulefill
                \begin{algorithmic}[1]
                    \STATE {\bfseries Function FindNeighbors}($X,\,\Mb{X}{\Wb}$) \hfill \% Proposition \ref{prp: FindNeighbors RSL_D}
                    \STATE $\Ne{X}{\Wb} \gets \Mb{X}{\Wb}$
                    \FOR{$Y \in \Mb{X}{\Wb}$}
                        \IF{$\exists Z\in \Mb{X}{\Wb}\setminus \{Y\}:\quad  \sep{X}{Y}{\Mb{X}{\Wb} \setminus \{Y,Z\}}{\D(\Wb)}$}
                            \STATE Remove $Y$ from $\Ne{X}{\Wb}$
                        \ENDIF
                    \ENDFOR
                    \STATE \textbf{return} $\Ne{X}{\Wb}$
                \end{algorithmic}
            \end{algorithm}
            
        \subsubsection{RSL\texorpdfstring{$_D$}{}}
            We present Algorithm \ref{alg: RSL_D} for RSL$_{D}$.
            The main body of the algorithm is the same as Algorithm \ref{alg: RSL_w} for RSL$_{\omega}$.
            However, the difference lies in the \textbf{IsRemovable} and \textbf{FindNeighbors} functions.
            The former uses Theorem \ref{thm: FindRemovable RSL_D} to check the removability of a variable, while the latter uses Proposition \ref{prp: FindNeighbors RSL_D} to learn the neighbors of a removable variable.
            Note that RSL$_D$ assumes that the underlying graph is diamond-free.

    \subsection{ROL} \label{subsec: implementation details - ROL}
        As we discussed in Section \ref{sec: ROL}, ROL aims to solve the optimization problem described in Equation \eqref{eq: opt problem}, which uses Algorithm \ref{alg: Gpi} to define a cost function for a given permutation.
        Additionally, we presented a reinforcement learning (RL) formulation to solve this optimization problem.
        In this section, we present the implementation details of three approaches for learning an r-order through solving Equation \eqref{eq: opt problem}.
        
        \subsubsection{ROL\texorpdfstring{$_{\text{HC}}$}{}}
            \begin{algorithm}[b!]
                \caption{ROL$_{\text{HC}}$}
                \label{alg: ROL_HC}
                \begin{algorithmic}[1]
                    \STATE \textbf{Input:} $\D(\Vb)$, \textsc{maxSwap}, \textsc{maxIter}
                    \STATE Initialize $\pi \in \Pi(\Vb)$
                    \STATE $C_{1n} \gets \textbf{ComputeCost}(\pi,1,n, \D(\Vb))$
                    \FOR{1 to \textsc{maxIter}}
                        \STATE Denote $\pi$ by $(X_1,\cdots, X_n)$
                        \FOR{$1 \leq a < b \leq n$ such that $b-a< \textsc{maxSwap}$} \label{line: ROL_HC}
                            \STATE $\pi_{\text{new}} \gets$ Swap $X_a$ and $X_b$ in $\pi$
                            \STATE $C^{\text{new}}_{ab} \gets \textbf{ComputeCost}(\pi_{\text{new}},a,b, \D(\Vb))$
                            \IF{$\sum_{i=a}^b C^{\text{new}}_{ab}(i) < \sum_{i=a}^b C_{1n}(i)$}
                                \STATE $\pi \gets \pi_{\text{new}}$
                                \FOR{$j$ from $a$ to $b$}
                                    \STATE $C_{1n}(j) \gets C^{\text{new}}_{ab}(j)$
                                \ENDFOR
                                \STATE \textbf{Break} the for loop of line \ref{line: ROL_HC}
                            \ENDIF
                        \ENDFOR
                    \ENDFOR
                    \STATE \textbf{Return} $\pi$
                \end{algorithmic}
                \hrulefill
                \begin{algorithmic}[1]
                    \STATE \textbf{Function ComputeCost} ($\pi,\,a,\,b,\, \D(\Vb)$)
                    \STATE $\Vb_a \gets \{X_a, X_{a+1},\dots, X_n\}$
                    \STATE $C_{ab} \gets (0,0,\cdots,0) \in \mathbb{R}^{n}$
                    \FOR{$t=a$ to $b$}
                        \STATE $X_t \gets \pi(t)$
                        \STATE $\Ne{X_t}{\Vb_t} \gets \textbf{FindNeighbors}(X_t, \D(\Vb_t))$
                        \STATE $C_{ab}(t) \gets |\Ne{X_t}{\Vb_t}|$
                        \STATE $\Vb_{t+1} \gets \Vb_t \setminus \{X_t\}$
                    \ENDFOR
                    \STATE \textbf{Return} $C_{ab}$
                \end{algorithmic}
            \end{algorithm}

            In Algorithm \ref{alg: ROL_HC}, we propose a hill-climbing approach, called ROL$_\text{HC}$ for finding an r-order.
            In general, the output of Algorithm \ref{alg: ROL_HC} is a suboptimal solution to \eqref{eq: opt problem} as it takes an initial order $\pi$ and gradually modifies it to another order with less cost, but it is not guaranteed to find a minimizer of \eqref{eq: opt problem} by taking such greedy approach.
            Nevertheless, this algorithm is suitable for practice as it is scalable to large graphs, and also achieves superior accuracy compared to the state-of-the-art methods.
    
            Inputs to Algorithm \ref{alg: ROL_HC} are the observational data $\D(\Vb)$ and two parameters \textsc{maxIter} and \textsc{maxSwap}.
            \textsc{maxIter} denotes the maximum number of iterations before the algorithm terminates, and \textsc{maxSwap} is an upper bound on the index difference of two variables that can get swapped in an iteration (line 6).
            Initial order $\pi$ in line 2 can be any arbitrary order, but selecting it cleverly (such as initialization using the output of other approaches) will improve the performance of the algorithm.

            The subroutine \textit{ComputeCost} takes as input an order $\pi$ and two numbers $1\leq a<b \leq n$ as input and returns a vector $C_{ab} \in \mathbb{R}^{n}$.
            For $a\leq t \leq b$, the $t$-th entry of $C_{ab}$ is equal to $|\Ne{X_t}{\Vb_t}|$ which is the number of neighbors of $X_t$ in the remaining graph.
            Hence, to learn the total cost of $\pi$, we can call this function with $a=1$ and $b=n$ and then compute the sum of the entries of the output $C_{1n}$.

            Accordingly, the main algorithm initially computes the cost vector of $\pi$ in line 3.
            The remainder of the algorithm (lines 4--13) updates $\pi$ iteratively, \textsc{maxIter} number of times.
            It updates the current order $\pi\!=\!(X_1,\!\cdots\!,X_n)$ as follows: first, it constructs a set of orders $\Pi^{\text{new}} \subseteq \Pi(\Vb)$ from $\pi$ by swapping any two variables $X_a$ and $X_b$ in $\pi$ as long as $1 \leq b-a \leq \textsc{maxSwap}$.
            Next, for each $\pi_{\text{new}} \in \Pi^{\text{new}}$, it computes the cost of $\pi_{\text{new}}$ and if it has a lower cost compared to the current order, the algorithm replaces $\pi$ by that order and repeats the process.
            Note that in line 8, the algorithm calls function \textit{ComputeCost} with $a$ and $b$ to compute the cost of the new policy.
            The reason is that for $i<a$ and $i>b$, the $i$-th entry of the cost of $\pi$ and $\pi_{\text{new}}$ are the same. This is because the set of remaining variables is the same.
            Hence, to compare the cost of $\pi$ with the cost of $\pi_{\text{new}}$, it suffices to compare them for entries between $a$ and $b$.
            Accordingly, Algorithm \ref{alg: ROL_HC} checks the condition in line 9, and if the cost of the new policy is better, then the algorithm updates $\pi$ and its corresponding cost.
            Note that it suffices to update the entries between $a$ to $b$ of $C_{1n}$.

        \subsubsection{ROL\texorpdfstring{$_{\text{VI}}$}{}} \label{sec: implementation - ROL_VI}
            Following the RL formulation of ROL introduced in Section \ref{sec: ROL}, we present ROL$_{\text{VI}}$.
            This algorithm uses value iteration to tackle the problem.
            
            Given a deterministic policy $\pi_\theta$ that is parameterized by $\theta$, we can adapt Algorithm \ref{alg: Gpi} to the RL setting as follows: the algorithm takes as input a policy $\pi_\theta$ instead of a permutation $\pi$.
            Furthermore, it uses the given policy to select $X_t$ in line 4, given by $X_t=\pi_\theta\left(\mathbf{V}_t\right)$.
            Finally, given a policy $\pi_\theta$, and the initial state $s_1=\Vb$, the cumulative reward of a trajectory $\tau=\left(s_1, a_1, s_2, a_2, \cdots, s_{n-1}, a_{n-1}\right)$, which denotes the sequence of states and actions selected by $\pi_\theta$, is given by
            \begin{align*}
                R\left(\tau_\theta\right)=\sum_{t=1}^{n-1} r\left(s_t, a_t\right)=-\sum_{t=1}^{n-1} \left|\Ne{a_t}{\G_{s_t}}\right|.
            \end{align*}
            Hence, if we denote the output of this modified algorithm by $\G^\theta=\left(\Vb, \Eb^\theta\right)$, then $R\left(\tau_\theta\right)=-\left|\Eb^\theta\right|$.
            With this RL formulation, any optimal policy-finding RL algorithm can be used to find a minimum-cost policy $\pi_\theta$ and thus solve the optimization problem given in Theorem \ref{thm: opt prob}.
            Accordingly, ROL$_{\text{VI}}$ applies value iteration algorithm.
            
        \subsubsection{ROL\texorpdfstring{$_{\text{PG}}$}{}} \label{sec: implementation - ROL_PG}
            Although any algorithm suited for RL is capable of finding an optimal deterministic policy for us, the complexity does not scale well as the graph size.
            Therefore, we can use a stochastic policy that increases the exploration during the training of an RL algorithm.
            ROL$_{\text{PG}}$ exploit stochastic policies parameterized by neural networks to further improve scalability.
            However, this could come at the price of approximating the optimal solution instead of finding the exact one.
            In the stochastic setting, an action $a_t$ is selected according to a distribution over the remaining variables, i.e., $a_t \sim P_{\theta}(\cdot\vert s_t=\Vb_t)$, where $\theta$ denotes the parameters of the policy (e.g., the weights used in training of a neural network).
            In this case, the objective of the algorithm is to minimize the expected total number of edges learned by policy $P_{\theta}(\cdot\vert s_t=\Vb_t)$, i.e.,
            \begin{equation}\label{eq: opt problem 2}
                \argmax_{\theta} \mathbb{E}_{\tau_{\theta} \sim P_{\theta}}\big[-|\Eb^{\theta}|\big],
            \end{equation}
            where the expectation is taken w.r.t. randomness of the stochastic policy.
            Many algorithms have been developed in the literature for finding stochastic policies and solving \eqref{eq: opt problem 2}.
            Some examples include Vanilla Policy Gradient (VPG) \citep{williams1992simple}, REINFORCE \citep{sutton1999policy},and Deep Q-Networks (DQN) \citep{mnih2013playing}.
            Accordingly, ROL$_{\text{VI}}$ applies VPG.

    \subsection{Initially Computing Markov Boundaries} \label{sec: ComputeMb function}
        Several algorithms have been proposed in the literature for discovering Markov boundaries \citep{margaritis1999bayesian, guyon2002gene, tsamardinos2003towards, tsamardinos2003algorithms, yaramakala2005speculative, fu2010markov}.
        One simple approach is to use total conditioning (TC) \citep{pellet2008using}.
        TC states that $X$ and $Y$ are in each other's Markov boundary if and only if
	\begin{equation*}
            \nsep{X}{Y}{\Vb \setminus \{X,Y\}}{P_{\Vb}}.
	\end{equation*}
        Using total conditioning, $\binom{n}{2}$ CI tests suffice to identify the Markov boundaries of all vertices.
        However, each test requires conditioning on a large set of variables.
        This issue has been addressed in multiple algorithms, including Grow-Shrink (GS) \citep{margaritis1999bayesian}, IAMB \citep{tsamardinos2003algorithms}, and its various modifications.
        These algorithms propose methods that perform more CI tests\footnote{These algorithms perform at most $\mathcal{O}(n^2)$ CI tests.} but with smaller conditioning sets.
        Choosing the right algorithm for computing the Markov boundaries depends on the available data.

    \subsection{Updating Markov Boundaries} \label{sec: UpdateMb}
        When a variable is removed in the recursive framework of Algorithm \ref{alg: Recursive framework}, we do not need to recompute the Markov boundaries of all the vertices.
        Instead, we can update the Markov boundaries of the remaining vertices. 

        Let $\Wb$ be the set of variables in an iteration with the set of Markov boundaries $\MB{\Wb}$.
        Suppose we want to remove a variable $X$ from $\Wb$ at the end of the current iteration, where $\Ne{X}{\Wb}$ is the set of neighbors of $X$.
        In this case, we only need to compute $\MB{\Wb \setminus \{X\}}$, which is the set of Markov boundaries of the remaining variables.
        
        \begin{algorithm}[t!]
            \caption{Updates Markov boundaries.}
            \label{alg: update Mb}
            \begin{algorithmic}[1]
                \STATE {\bfseries UpdateMb}($X,\,\Ne{X}{\Wb},\,\MB{\Wb}$)
                \STATE $\MB{\Wb \setminus \{X\}} \gets \{\Mb{Y}{\Wb}\!:\: Y \in \Wb \setminus \{X\}\}$ 
                \FOR{$Y\in \Mb{X}{\Wb}$}
                    \STATE Remove $X$ from $\Mb{Y}{\Wb \setminus \{X\}}$
                \ENDFOR
                    \FOR{$Y,Z \in \Ne{X}{\Wb}$ such that $|\Mb{Y}{\Wb}| \leq |\Mb{Z}{\Wb}|$} 
                        \IF{$\sep{Y}{Z}{\Mb{Y}{\Wb \setminus \{X\}} \setminus \{Y,Z\}}{\D(\Wb)}$}
                            \STATE Remove $Z$ from $\Mb{Y}{\Wb \setminus \{X\}}$ 
                            \STATE Remove $Y$ from $\Mb{Z}{\Wb \setminus \{X\}}$
                            \STATE $\F(Y) \gets$ \false
                            \STATE $\F(Z) \gets$ \false
                        \ENDIF
                    \ENDFOR
                \RETURN $\MB{\Wb \setminus \{X\}}$
            \end{algorithmic}
        \end{algorithm}

        We can use Algorithm \ref{alg: update Mb} to compute $\MB{\Wb \setminus \{X\}}$.
        Removing vertex $X$ from MAG $\G_{\Wb}$ has two effects.
        \begin{enumerate}
            \item $X$ is removed from all Markov boundaries, and
            \item for $Y,Z \in \Wb \setminus \{X\}$, if all of the collider paths between $Y$ and $Z$ in $\G_{\Wb}$ pass through $X$, then $Y$ and $Z$ must be excluded from each others Markov boundary.
        \end{enumerate}
        In the latter case, $Y, Z \in \Mb{X}{\Wb}$ and the update is performed using a single CI test,
        \begin{equation*}
            \nsep{Y}{Z}{\Mb{Z}{\Wb} \setminus \{X, Y, Z\}}{P_{\Wb}}, \quad \text{or equivalently}, \quad \nsep{Y}{Z}{\Mb{Y}{\Wb} \setminus \{X, Y, Z\}}{P_{\Wb}}.
        \end{equation*}
        We chose the CI test with the smaller conditioning set.
        If the CI test shows that $Y$ and $Z$ are conditionally independent, we remove them from each other's Markov boundary.

        Recall that we used \F in the proposed algorithms to avoid unnecessary computations.
        As discussed in Section \ref{sec: duplicate CI tests MARVEL}, when the Markov boundary of a variable has changed, we need to set its \F value to be \false.
        Accordingly, when the If condition of line 6 holds, Algorithm \ref{alg: update Mb} removes $Y$ and $Z$ from each other's Markov boundary.
        Therefore, we set their \F values to \false in lines 9 and 10.
    
    \subsection{Identifying the MEC} \label{sec: MEC}
        In the previous sections, we primarily addressed the task of learning the skeleton of the causal graph.
        In this section, we discuss how our methods can be augmented to identify the MEC of the underlying causal mechanisms ($[\GV]$).
        As a general rule, having access to the true skeleton and a separating set for each pair of non-neighbor vertices suffice to identify the MEC
        \citep{zhang2008completeness}.
        However, we will provide modifications tailored to a few of our algorithms, which will improve computational complexity.
        
        Algorithms MARVEL and RSL require causal sufficiency.
        This implies that their objective is to recover the Markov equivalence class of a DAG, $[\G]^d$.
        \citet{verma1991equivalence} showed that two DAGs are Markov equivalent if and only if they have the same skeleton and v-structures.
        Accordingly, to identify $[\G]^d$, it suffices to learn the skeleton and v-structures of DAG $\G$.
        As such, in Sections \ref{sec:mecmarvel} and \ref{sec:mecrsl} we describe how to recover the v-structures.
        With this information at hand, Meek rules \citep{meek1995causal} can be applied to achieve a \emph{maximally oriented} DAG (CPDAG), also known as \emph{essential graph}.
        For characterization and graph-theoretical properties of such graphs, refer to \cite{andersson1997characterization}.
        
        On the other hand, Algorithms L-MARVEL and ROL serve to recover the MEC of a MAG.
        Having the same skeleton and unshielded colliders is necessary but not sufficient for two MAGs to be Markov equivalent.
        The following proposition by \cite{spirtes1996polynomial} characterizes necessary and sufficient conditions for two MAGs to be Markov equivalent.
        \begin{proposition}[\citealp{spirtes1996polynomial}]
            Two MAGs $\G_1$ and $\G_2$ are Markov equivalent if and only if (i) they have the same skeleton, (ii) they have the same unshielded colliders, and (iii) if a path $\mathcal{P}$ is a discriminating path (Definition \ref{def:discpath}) for a vertex $X$ in both MAGs, then $X$ is a collider on $\mathcal{P}$ in $\G_1$ if and only if it is a collider on $\mathcal{P}$ in $\G_2$.
        \end{proposition}
        Building upon this proposition, it suffices to learn the skeleton, unshielded colliders, and shielded colliders for which a discriminating path exists.
        Subsequently, complete orientation rules can be applied to achieve a maximally oriented (aka maximally informative) \emph{partial ancestral graph} (PAG).
        We refer to \citet{zhang2008completeness} for complete orientation rules and further discussion.

        \subsubsection{Recover V-structures in MARVEL} \label{sec:mecmarvel}
            Our goal is to recover the v-structures.
            Note that a v-structure comprises a pair of co-parents and a common child of them.
            As such, we first describe how to identify the pairs of variables that are co-parents.
            
            Since each variable is identified as removable in exactly one iteration of MARVEL, we can identify the co-parents of a variable at the iteration where it gets removed.
            However, removing a variable does not preserve co-parent relationships.
            Indeed, if $Y,Z$ are co-parents in a DAG $\G$, where $X$ is a removable variable, then either $Y,Z$ are still co-parents of each other in $\G[\Vb \setminus \{X\}]$, or $Y\in \Mb{Z}{\G} \setminus \Mb{Z}{\G[\Vb \setminus \{X\}]}$.
            The latter case happens when $X$ is the only common child of $Y$ and $Z$.
            Based on this observation, in order to identify all co-parent pairs,
            we first modify Algorithm \ref{alg: update Mb}, the procedure for updating Markov boundaries, so that whenever a pair of variables $(Y, Z)$ are removed from each other's Markov boundary in lines 7 and 8, this pair is marked as co-parents of each other in the final graph.
            Furthermore, the separating set for this pair, namely $\Mb{Y}{\Wb\setminus\{X\}}$, is stored as $\Sb_{YZ}$. 
            The rest of the co-parents and their corresponding separating sets are discovered in line 13 of Algorithm \ref{alg: MARVEL}, which consists of an application of Lemma \ref{lem: neighbor}.
            During this step, the Markov boundary of $X$ is partitioned into $\MB{\mathbf{V}_i}(X)=\Ne{X}{\Vb_i}\sqcup\CP{X}{\Vb_i}$ through finding a separating set $\Sb_{XY}$ for every variable $Y\in\CP{X}{\Vb_i}$.
            These separating sets are stored for every pair of co-parents $(X, Y)$.
            At the end of the skeleton discovery phase, the set of v-structures can be identified based on the following lemma.
            
            \begin{lemma}[\citealp{verma1991equivalence}]\label{lem:v-structures}
                Let $X, Y, Z$ be three arbitrary vertices of DAG $\G$.
                These vertices form a v-structure in $\G$, i.e., $X\to Z\gets Y$ if and only if all of the following hold:
                \begin{equation*}
                    Y\in\CP{X}{\G}, \quad Z\in\Ne{X}{\G}\cap\Ne{Y}{\G}, \quad  Z\notin \Sb_{XY}.
                \end{equation*}
            \end{lemma}
            Accordingly, Lemma \ref{lem:v-structures} can be integrated into MARVEL to identify the v-structures.
            


            \subsubsection{Recover V-structures in RSL}\label{sec:mecrsl}
                In analogy to MARVEL, it suffices to identify the v-structures.
                Also, the v-structures that are not preserved due to vertex removals can be identified and oriented through modifying Algorithm \ref{alg: update Mb}, just as described in the case of MARVEL.
                Herein, we present the procedure for identifying the v-structures $\VS{X}{\G}$ for a removable variable $X$.
                We describe this procedure for RSL$_D$ and RSL$_\omega$ separately.
                
                \textbf{RSL\texorpdfstring{$_D$}{}.}
                In the case of diamond-free graphs, if $X$ is a removable variable and $Y\in\CP{X}{\G}$, then $X$ and $Y$ have a \emph{unique} common child.
                Indeed, Proposition \ref{prp: FindNeighbors RSL_D} reveals not only the co-parents of a removable variable $X$ but also the unique common child of $X$ and each of its co-parents.
                Accordingly, to identify the v-structures $\VS{X}{\G}$ during RSL$_D$, 
                it suffices to modify the \textbf{FindNeighbors} subroutine as follows.
                Each time a variable $Z\in \Mb{X}{\G}\setminus \{Y\}$ satisfies Equation \eqref{eq:v-rsld} in line 4 of \textbf{FindNeighbors}, the edges are oriented as $X\to Z$ and $Y\to Z$, since $Z$ is the unique common child of $X$ and $Y$.

                \textbf{RSL\texorpdfstring{$_\omega$}{}.}
                Analogous to RSL$_D$, we can exploit the procedure for finding the neighbors to further identify the v-structures.
                In particular, Proposition \ref{prp: FindNeighbors clique} identifies a unique set $\Sb$ as the common children of $X$ and $Y$, for any $Y$ that is a co-parent of $X$.
                Once such a set $\Sb$ is found in line 4 of \textbf{FindNeighbors} in RSL$_\omega$ (see line 12 of Algorithm \ref{alg: RSL_w}) for a removable variable $X_j$ and $Y\in\Mb{X_j}{\G}$, it suffices to orient every edge $U-V$ such that $U\in\{X_j, Y\}$ and $V\in\Sb$ as $U\to V$.

\section{Complexity and Completeness Analysis} \label{sec: complexity}
    In this section, we discuss the complexity and completeness of various causal discovery methods, with a particular emphasis on our proposed recursive approaches.
    This analysis is crucial for understanding the efficiency and reliability of these methods in practical applications.
    In addition, we delve into the theoretical limits of these algorithms by providing lower bounds for the complexity of constraint-based algorithms in learning DAGs and MAGs.

    \subsection{Lower Bound} \label{sec: lower bound}
        We introduce two fundamental theorems: Theorem \ref{thm: lower Bound DAG}, which establishes a lower bound for the complexity of constraint-based algorithms in learning DAGs, and Theorem \ref{thm: lower Bound MAG}, which does the same for MAGs.
        These theorems are instrumental in quantifying the theoretical limits of constraint-based methods.
    
        \begin{theorem}[Lower bound for DAGs] \label{thm: lower Bound DAG}
            For any positive integers $n$ and $1 \leq c \leq n$, there exists a DAG $\G$ with $n$ vertices and $\Delta_{in}(\G) = c$ such that the number of d-separations of the form $\sep{X}{Y}{\Zb}{\G}$ required by any constraint-based algorithm to learn the skeleton of $\G$ is lower bounded by
            \begin{equation} \label{eq: lower bound DAG}
              \Omega(n^2 + n\Delta_{in}(\G) 2^{\Delta_{in}(\G)}).
            \end{equation}
        \end{theorem}
        Theorem \ref{thm: lower Bound DAG} determines the hardness of causal discovery under causal sufficiency parameterized based on $\Delta_{in}(\G)$ rather than other graph parameters such as $\Delta(\G)$ or $\alpha(\G)$.

        \begin{theorem}[Lower bound for MAGs] \label{thm: lower Bound MAG}
            For any positive integers $n$ and $1 \leq c \leq n$, there exists a MAG $\G$ with $n$ vertices and $\Delta_{in}^+(\G) = c$ such that the number of m-separations of the form $\sep{X}{Y}{\Zb}{\G}$ required by any constraint-based algorithm to learn the skeleton of $\G$ is lower bounded by
            \begin{equation} \label{eq: lower bound MAG}
              \Omega(n^2 + n\Delta_{in}^+(\G) 2^{\Delta_{in}^+(\G)}).
            \end{equation}
        \end{theorem}
        Theorem \ref{thm: lower Bound MAG} extends the complexity analysis to the setting of MAGs, addressing scenarios in causal discovery where the assumption of causal sufficiency does not hold.
        It similarly parameterizes the hardness of the problem based on $\Delta_{in}^+(\G)$, which is an extension of $\Delta_{in}$ for MAGs (Definition \ref{def: pap}).

    \subsection{Completeness Analysis and Achievable Bounds of RCD Methods}
        We present completeness guarantees and upper bounds on the number of performed CI tests by MARVEL, L-MARVEL, RSL, and ROL, as implemented in Section \ref{sec: implementation details}.
        
        Recall that our algorithms take as input $\D(\Vb)$, a set of i.i.d. samples from the observational distribution $\PV$.
        In the propositions that follow, we provide asymptotic guarantees of correctness for our methods under the assumption that $\D(\Vb)$ is sufficiently large to accurately recover the CI relations present in $\PV$.
        Additionally, as outlined in Section \ref{sec: preliminaries}, we assume that the true underlying graph encodes the CI relations of the observational distribution, as stated in Equation \eqref{eq: PV iff GV}.
        It should be noted that while our methods are also compatible with weaker notions of faithfulness, we focus on this primary assumption for simplicity in our presentation.

        \begin{proposition}[Completeness and complexity of MARVEL] \label{prp: complexity of MARVEL}
            Under the assumption of causal sufficiency, MARVEL, as implemented in Algorithm \ref{alg: MARVEL}, correctly returns the skeleton of DAG $\G$ by performing at most
		\begin{equation} \label{eq: MARVEL upper bound}
                \binom{n}{2} + n \binom{\Delta_{in}(\G)}{2} + \frac{n}{2}\Delta_{in}(\G)(1+ 0.45\Delta_{in}(\G) )2^{\Delta_{in}(\G)}
                = \mathcal{O}\big(n^2 + n \Delta_{in}(\G)^2 2^{\Delta_{in}(\G)}\big)
		\end{equation}
            unique CI tests, and at most $\binom{n}{2}2^{\Delta_{in}(\G)-1}$ duplicate CI tests in the worst case\footnote{Duplicate CI tests can be completely eliminated through efficient use of memory. This aspect is omitted here for simplicity and readability.}.
	\end{proposition}
        \begin{corollary}
            If $\Delta_\text{in}(\G) \leq c \log n$, MARVEL uses at most $\mathcal{O}(n^2 + n^{c+1}\log^2 n)$ unique CI tests in the worst case, which is polynomial in the number of variables.
        \end{corollary}

        As $n$ gets larger, if DAG $\G$ has a constant value of $\Delta_{in}(\G)$, or more generally $\Delta_{in}(\G) \leq (1-\epsilon)\log n$, where $\epsilon>0$, then both the achievable upper bound of MARVEL in \eqref{eq: MARVEL upper bound} and the lower bound in \eqref{eq: lower bound DAG} are quadratic in $n$.
        For larger values of $\Delta_{in}(\G)$, the second terms in these equations become dominant.
        In this case, the upper bound of MARVEL differs from the lower bound only by a factor of $\Delta_{in}(\G)$.
        This demonstrates that under causal sufficiency and without any additional information, MARVEL has a worst-case performance that nearly matches the lower bound.

        \begin{proposition}[Completeness and complexity of L-MARVEL] \label{prp: complexity of L-MARVEL}
            L-MARVEL, as implemented in Algorithm \ref{alg: L-MARVEL}, correctly returns the skeleton of MAG $\GV$ by performing at most
            \begin{equation*}
                \mathcal{O}\big(n^2 + n \Delta_{in}^+(\GV)^2 2^{\Delta_{in}^+(\GV)}\big)
            \end{equation*}
            number of CI tests.
        \end{proposition}
        Similar to our previous argument for MARVEL, Proposition \ref{prp: complexity of L-MARVEL} shows that the complexity of L-MARVEL aligns closely with the lower bound for MAGs in Theorem \ref{thm: lower Bound MAG}, demonstrating its near-optimal performance in causal models with latent variables.

        \begin{proposition}[Completeness and complexity of RSL$_\omega$] \label{prp: complexity of RSL_w}
            Under causal sufficiency, if $\omega(\G)\leq m$, then RSL$_\omega$, as implemented in Algorithm \ref{alg: RSL_w}, correctly returns the skeleton of DAG $\G$ by performing at most
		\begin{equation*}
                \mathcal{O}\big(n^2 + n\Delta_{in}(\G)^{m+1}\big)
		\end{equation*}
            number of CI tests.
	\end{proposition}
        Proposition \ref{prp: complexity of RSL_w} marks a pivotal development in causal discovery for DAGs with bounded \emph{TreeWidth}.
        Recent studies including \cite{korhonen2013exact, nie2014advances, ramaswamy2021turbocharging}, have emphasized the importance of algorithms tailored for scenarios where an upper bound on the TreeWidth of the causal graph is given as side information.
        A bound on TreeWidth is more restrictive than a bounded clique number, as indicated by the following inequality.
        \begin{equation*}
            \omega(\G) \leq \text{TreeWidth}(\G) + 1
        \end{equation*}
        As such, our RSL$_\omega$ algorithm is also applicable to causal discovery in DAGs with bounded TreeWidth.
        Notably, while existing exact discovery algorithms for these networks demonstrate exponential complexity, RSL$_\omega$ maintains a polynomial complexity.
        This indicates that when the TreeWidth is constant, causal discovery is no longer NP-hard and can be solved in polynomial time.

        \begin{proposition}[Completeness and complexity of RSL$_D$] \label{prp: complexity of RSL_D}
            Under causal sufficiency and if $\G$ is a diamond-free DAG, then RSL$_D$, as implemented in Algorithm \ref{alg: RSL_D}, correctly returns the skeleton of DAG $\G$ by performing at most
		\begin{equation} \label{eq: upper bound - RSL_D}
                \mathcal{O}\big(n^2 + n\Delta_{in}(\G)^3\big)
		\end{equation}
            number of CI tests.
	\end{proposition}
        \begin{remark}
            Even if DAG $\G$ has diamonds, RSL$_D$ correctly recovers all the existing edges with possibly extra edges, i.e., RSL$_D$ has no false negative.
        \end{remark}
        RSL$_D$ is the fastest among our proposed recursive methods.
        This can also be seen by the upper bound in Equation \eqref{eq: upper bound - RSL_D}, where the number of CI tests stays tractable for graphs with more than 1000 variables.
        
        \begin{proposition}[Complexity of ROL$_{\text{HC}}$] \label{prp: complexity of ROL_HC}
            ROL$_{\text{HC}}$, as implemented in Algorithm \ref{alg: ROL_HC}, performs at most
            \begin{equation} \label{eq: complexity ROL_HC}
                \mathcal{O}\big(\textsc{maxIter} \times n^3\big)
            \end{equation}
            number of CI tests, excluding the initialization step in line 2.
        \end{proposition}
        Note that the upper bound in Equation \eqref{eq: complexity ROL_HC} may vary depending on the initialization step in line 2 and the choice of the \textit{FindNeighbors} function.
        Also, here we are assuming that \textsc{maxSwap} is a constant.
        
        \begin{proposition}[Completeness and complexity of ROL$_{\text{VI}}$] \label{prp: complexity of ROL_VI}
            According to the introduced RL setting in Section \ref{sec: implementation - ROL_VI},  ROL$_{\text{VI}}$ finds the optimal policy by performing at most $\mathcal{O}(n^2 2^n)$ number of CI tests.
        \end{proposition}
        Note that the bound in Proposition \ref{prp: complexity of ROL_VI} is much lower than $\mathcal{O}(n!)$ for iterating over all orders.

    \subsection{Comparison}
        \begin{table}[t!]
            \fontsize{9}{10.5}\selectfont
            \centering
            \begin{tabular}{N M{2cm}|M{2.7cm} M{2cm}|M{2.4cm}| M{4.2cm}}
                \toprule
                &\multirow{2}{*}{\textbf{Algorithm}} & \multicolumn{2}{c|}{\textbf{Assumptions}}  & \multirow{2}{*}{\textbf{Completeness}}& \multirow{2}{*}{\textbf{\#CI tests}}
                \\
                & 
                & Causal sufficiency
                & Other
                & 
                & 
                \\
                \hline
                & MARVEL
                & YES
                & -
                & YES
                & $\mathcal{O}(n^2 + n\Delta_{in}^2 2^{\Delta_{in}})$
                \\
                & L-MARVEL
                & NO
                & -
                & YES
                & $\mathcal{O}(n^2 + n ({\Delta_{in}^+})^2 2^{\Delta_{in}^+})$
                \\
                & RSL$_\omega$
                & YES
                & $\omega(\G) \leq m$
                & YES
                & $\mathcal{O}(n^2 + n\Delta_{in}^{m+1})$
                \\
                & RSL$_D$
                & YES
                & Diamond-free
                & YES
                & $\mathcal{O}(n^2 + n\Delta_{in}^3)$
                \\
                & ROL$_{\text{HC}}$
                & NO
                & -
                & NO
                & $\mathcal{O}(\textsc{maxIter} \times n^3)$
                \\
                & ROL$_{\text{VI}}$
                & NO
                & -
                & YES
                & $\mathcal{O}(n^2 2^n)$
                \\
                & ROL$_{\text{PG}}$
                & NO
                & -
                & NO
                & N/A
                \\
                \hline
                & PC
                & YES
                & -
                & YES
                & $\mathcal{O}(n^\Delta)$
                \\
                & GS
                & YES
                & -
                & YES
                & $\mathcal{O}(n^2+ n \alpha^2 2^\alpha)$
                \\
                & MMPC
                & YES
                & -
                & YES
                & $\mathcal{O}(n^22^\alpha$)
                \\
                & CS
                & YES
                & -
                & YES
                & $\mathcal{O}(n^22^\alpha$)
                \\
                & FCI
                & NO
                & -
                & YES
                & N/A
                \\
                & RFCI
                & NO
                & -
                & NO
                & N/A
                \\
                & FCI+
                & NO
                & -
                & NO
                & $\mathcal{O}(n^{2(\Delta+2)})$
                \\
                & MBCS*
                & NO
                & -
                & YES
                & $\mathcal{O}(n^22^\alpha$)
                \\
                \hline
                & Lower Bound
                & YES
                & -
                & YES
                & $\Omega(n^2 + n\Delta_{in} 2^{\Delta_{in}})$
                \\
                & Lower Bound
                & NO
                & -
                & YES
                & $\mathcal{O}(n^2 + n {\Delta_{in}^+} 2^{\Delta_{in}^+})$
                \\
                \bottomrule
            \end{tabular}
            \caption{Summary of the assumptions, guarantees, and the complexity of various causal discovery methods from observational data.}
            \label{table: all algorithms}
        \end{table}

        In this part, we present a comparative analysis of various causal discovery methods, including our proposed algorithms.
        The summary of this comparison is presented in Table \ref{table: all algorithms}, which categorizes each algorithm based on its assumptions, completeness guarantees, and computational complexity in terms of the number of CI tests required.
        The last two rows present the lower bounds for complexity under causal sufficiency and in the absence of it, as established in Section \ref{sec: lower bound}.

        A critical observation from this analysis is that the upper bounds on the complexity of our algorithms (MARVEL, L-MARVEL, RSL, and ROL) are significantly more efficient compared to the others.
        This efficiency in DAGs is largely due to the following inequality.
        \begin{equation*}
            \Delta_{in}(\G) \leq \Delta(\G) \leq \alpha(\G)
            \label{eq: complexity-inequality}
        \end{equation*}
        Additionally, in a DAG with a constant in-degree, $\Delta$ and $\alpha$ can grow linearly with the number of variables.

        \begin{corollary}
            Under the assumption of causal sufficiency, RSL$_D$ is the fastest among our proposed methods.
            In scenarios lacking causal sufficiency, L-MARVEL is fastest for sparse graphs, while ROL$_{\text{HC}}$ outperforms in denser graphs.
        \end{corollary}

\section{Related Work} \label{sec: related work}
    Causal discovery methods can be broadly categorized into constraint-based and score-based approaches.
    While the former group recovers the structure consistent with the conditional independence constraints, the latter opts for the graph that maximizes a specific score function.
    In this section, we present an overview of relevant works in the field, organizing them into these two main categories.
    Under the score-based category, we give special consideration to the so-called \emph{ordering-based} methods, which rely on recovering a specific order among the variables to guide the causal discovery task.
    The papers presented in this section are summarized in Table \ref{table:related}.
    While we strive to provide a representative review of key works in causal discovery, we note that the vast body of literature includes many more studies and approaches.
    For a more comprehensive compilation and deeper discussion, we refer to surveys on causal discovery such as \cite{vowels2022d}, \cite{glymour2019review}, and \cite{mooij2016distinguishing}.

    \begin{table}[p]
        \fontsize{9}{10.5}\selectfont
        \centering
        \begin{adjustbox}{width=0.98\textwidth, totalheight=0.95\textheight, keepaspectratio}
            \begin{tabular}{N M{2.1cm}|M{1.8cm}|M{2.3cm}|M{2.2cm}| M{4.1cm}}
                \toprule
                &\multirow{2}{*}{\textbf{Algorithm}} & \multirow{2}{*}{\textbf{Type}} & 
                \multirow{2}{*}{\textbf{\shortstack{Causal \\Sufficiency}}}
                & \multirow{2}{*}{\hspace{-1pt}\textbf{Completeness}}& \multirow{2}{*}{\textbf{Reference}}
                \\
                & 
                & 
                &
                & 
                & 
                \\
                \hline
                & PC 
                & Constraint
                & YES
                & YES
                & \cite{spirtes2000causation}
                \\
                & GS 
                & Constraint
                & YES
                & YES
                & \cite{margaritis1999bayesian}
                \\
                & CPC 
                & Constraint
                & YES
                & YES
                & \cite{ramsey2006adjacency}
                \\
                & KCL 
                & Constraint
                & YES
                & YES
                & \cite{sun2007kernel}
                \\
                & TC
                & Constraint
                & YES
                & YES
                & \scriptsize\cite{pellet2008using}
                \\
                & KCI-test 
                & Constraint
                & YES
                & YES
                & \cite{zhang2011kernel}
                \\
                & RPC
                & Constraint
                & YES
                & YES
                & \cite{harris2013pc}
                \\
                & PC-stable
                & Constraint
                & YES
                & YES
                & \cite{colombo2014order}
                \\
                & Parallel-PC
                & Constraint
                & YES
                & YES
                & \cite{le2016fast}
                \\

                & HGC
                & Score
                & YES
                & YES
                & \cite{heckerman1995learning}
                \\
                & GES
                & Score
                & YES
                & YES
                & \cite{chickering2002optimal}
                \\
                & HGL
                & Score
                & YES
                & YES
                & \cite{he2005learning}
                \\
                & CD
                & Score
                & YES
                & NO
                & \cite{fu2013learning}
                \\
                & CAM
                & Score
                & YES
                & YES
                & \cite{buhlmann2014cam}
                \\
                & PNL
                & Score
                & YES
                & YES
                & \cite{aragam2015learning}
                \\
                & FGES
                & Score
                & YES
                & YES
                & \cite{ramsey2017million}
                \\

                & NO TEARS
                & Score
                & YES
                & YES
                & \cite{zheng2018dags}
                \\
                & CGNN
                & Score
                & YES
                & YES
                & \cite{goudet2018learning}
                \\
                & DAG-GNN
                & Score
                & YES
                & YES
                & \cite{yu2019dag}
                \\
                & Graph AutoEncoder
                & Score
                & YES
                & YES
                & \cite{ng2019graph}
                \\
                & NO BEARS
                & Score
                & YES
                & YES
                & \cite{lee2019scaling}
                \\
                & DYNOTEARS
                & Score
                & YES
                & YES
                & \cite{pamfil2020dynotears}
                \\
                & SAM
                & Score
                & YES
                & YES
                & \cite{kalainathan2022structural}
                \\
                & OS
                & Ordering
                & YES
                & NO
                & \cite{teyssier2005ordering}
                \\
                & L1-RP
                & Ordering
                & YES
                & YES
                & \cite{schmidt2007learning}
                \\
                & ASOBS
                & Ordering
                & YES
                & NO
                & \cite{scanagatta2015learning}
                \\
                & SP
                & Ordering
                & YES
                & YES
                & \cite{raskutti2018learning}
                \\
                & TSP
                & Ordering
                & YES
                & YES
                & \cite{solus2021consistency}
                \\
                & GRaSP
                & Ordering
                & YES
                & YES
                & \cite{lam2022greedy}
                \\

                & FCI 
                & Constraint
                & NO
                & YES
                & \cite{spirtes1995causal}
                \\
                & ION 
                & Constraint
                & NO
                & YES
                & \cite{danks2008integrating}
                \\
                & MBCS*
                & Constraint
                & NO
                & YES
                & \scriptsize\cite{pellet2008finding}
                \\
                & RFCI 
                & Constraint
                & NO
                & NO
                & \cite{colombo2012learning}
                \\
                & GSPo
                & Ordering
                & NO
                & NO
                & \cite{bernstein2020ordering}
                \\
                \bottomrule
            \end{tabular}
        \end{adjustbox}
        \caption{Summary of various causal discovery methods from observational data.}
        \label{table:related}
    \end{table}

    \subsection{Constraint-Based}
        PC algorithm \citep{spirtes2000causation}, widely used for causal discovery, stands as a foundational approach for this task using observational data, laying the groundwork for most of the subsequent developments in constraint-based approaches.
        Notably, acknowledging the high computational cost of PC, \citet{le2016fast} provided a parallel computing framework for it.
        \citet{colombo2014order} drew attention to the fact that under potentially erroneous conditional independence tests, the results obtained by the PC algorithm may depend on the order in which these tests were conducted.
        Following this observation, they introduced PC-stable, which offers an order-independent approach to causal discovery.
        This enhances the PC's robustness with respect to uncertainties over the order of variables.
        Further, RPC \citep{harris2013pc} was introduced as a relaxation of the PC algorithm to handle instances where strict adherence to conditional independence tests may not be possible.
        
        PC and its derivatives work under causal sufficiency.
        In extending the scope of constraint-based methods, \cite{spirtes1995causal} introduced the fast causal inference algorithm (FCI), which accommodates latent variables and selection bias.
        Although foundational, FCI faced challenges with incomplete edge orientation rules.
        \citet{zhang2008completeness} later augmented further orientation rules to ensure completeness of its output.
        FCI also suffers from an intractable computational complexity in moderate to high dimensional causal discovery tasks.
        Focusing on learning high-dimensional causal graphs, \citet{colombo2012learning} introduced RFCI, which performs only a subset of the conditional independence tests that FCI requires.
        Faster computations came at the cost of not being \emph{complete}, in the sense that the recovered graph may contain extra edges in general.
        
        Closest to our approach are the works by \citet{margaritis1999bayesian}, \citet{pellet2008finding}, and \citet{pellet2008using}, which put forward the idea of using Markov boundary information to guide causal discovery.
        The grow-shrink (GS) algorithm \citep{margaritis1999bayesian} was originally devised as a method to infer Markov boundaries.
        The authors augmented GS with further steps to make it capable of recovering the causal structure.
        \citet{pellet2008using} proposed the use of total conditioning (TC) to infer the Markov boundary information and recover the causal structure afterward.
        \citet{pellet2008finding} also generalized the same ideas to handle causal discovery in the presence of latent variables and selection bias.
        
        Along a separate path, CPC \citep{ramsey2006adjacency} introduced the concept of adjacency-faithfulness, building a conservative framework for constraint-based causal discovery.
        KCL \citep{sun2007kernel} presented a kernel-based causal discovery algorithm, extending the scope of applicability of causal discovery algorithms to non-linearly related structural models through the use of kernel methods.
        On the same note, the kernel-based conditional independence test (KCI-test) introduced by \citet{zhang2008causal} further enriches the toolkit for assessing conditional independence through kernel methods.
        ION \citep{danks2008integrating} focused on integrating observational data with narrative information to refine causal relationships.
        
    \subsection{Score-Based}
        Score-based methods provide an alternative approach to constraint-based methods, emphasizing the optimization of a score function to identify the most likely causal graph.
        The foundation for this line of research was laid by early work in Bayesian statistics, as well as the development of graphical models \citep{pearl1988probabilistic}.
        A notable contribution was made by \citet{heckerman1995learning}, which put forward the idea of integrating the prior beliefs and statistical data through a Bayesian approach to causal discovery.
        The authors reviewed certain heuristic algorithms to search for the graphical structure maximizing their scoring function.
        Greedy equivalence search (GES) algorithm, introduced by \citet{chickering2002optimal}, represents another significant advancement in the field.
        GES employs a step-wise greedy search strategy to iteratively refine the graph structure, aiming for maximizing the Bayesian information criterion \citep{raftery1995bayesian, geiger1994learning}.
        
        GES has been influential in guiding the subsequent advancements in score-based causal discovery.
        An extension was introduced by \citet{buhlmann2014cam}, which diverges from GES by decoupling the order search among variables and edge selection in the DAG from each other.
        While the variable order search is carried out through a non-regularized maximum likelihood estimation, sparse regression techniques are used for edge selection.
        The method developed by \citet{buhlmann2014cam} is valid for additive models.
        Another extension is the fast greedy equivalence search (FGES) \citep{ramsey2017million}.
        FGES builds upon GES by introducing two modifications to increase the speed of the search in order to adapt it for high-dimensional causal models.
        
        Beyond greedy search approaches, there is also literature based on coordinate descent for optimizing the score function.
        \citet{fu2013learning} utilizes an $\ell_1$-regularized likelihood approach and a block-wise coordinate descent to estimate the causal structure.
        \citet{gu2019penalized} model the conditional density of a variable given its parents by multi-logit regression, employing a group norm penalty to obtain a sparse graph.
        \citet{aragam2015learning} reduce causal discovery to a series of neighborhood regressions under suitable assumptions.
        
        A broad range of recent research on score-based causal discovery has focused on methods based on continuous optimization.
        The most influential work in this direction was DAGs with NO TEARS \citep{zheng2018dags}, reformulating the combinatorial problem of causal search into a continuous optimization problem.
        \citet{goudet2018learning} introduced CGNN, which uses neural networks to learn the functional mappings between variables and incorporates a hill-climbing search algorithm for the optimization.
        In order to address the computational costs of CGNN, \citet{kalainathan2022structural} presented SAM.
        SAM avoids the computational complexity of CGNN through the use of adversarial training, and end-to-end training of the SAM architecture and weights.
        Other notable extensions of NO TEARS include but are not limited to DAG-GNN \citep{yu2019dag}, Graph AutoEncoder \citep{ng2019graph}, NO BEARS \citep{lee2019scaling}, and DYNOTEARS \citep{pamfil2020dynotears}.
        
    \subsection{Ordering-based}
        Despite the inherent difficulty of causal discovery \citep{chickering1996learning}, finding the graphical structure that maximizes a scoring function becomes tractable when an ordering is postulated on the variables \citep{buntine1991theory, cooper1992bayesian}.
        Based on this observation, \citet{teyssier2005ordering} proposed a search over the space of variable orderings, rather than the previously adopted search over the space of DAGs.
        After recovering the ordering among variables, \citet{teyssier2005ordering} used an exhaustive search through all possible parent sets for each vertex.
        Improving on the latter, \citet{schmidt2007learning} showed that this search can be well-approximated through $\ell_1$ regularization.
        Another extension to the work by \citet{teyssier2005ordering} was introduced by \citet{scanagatta2015learning}, who proposed anytime algorithms to circumvent the high costs of searching in the space of potential parent sets.
        \citet{raskutti2018learning} introduced the sparsest permutation (SP) algorithm, which relaxes the common faithfulness assumption to a weaker assumption called \emph{u-frugality}.
        On the same note, \citet{lam2022greedy} developed a class of permutation-based algorithms, namely GRaSP, which operate under weaker assumptions than faithfulness.
        \citet{solus2021consistency} were the first to provide consistency guarantees for a greedy permutation-based search algorithm, namely GSP.
        \citet{bernstein2020ordering} extended the scope of permutation-based methods to causal structures with latent variables.
        
    \subsection{Miscellaneous}
        While the focus of this paper is on methods for learning a causal graph based on observational data, it is essential to acknowledge other directions in causal discovery research.
        For example, \cite{yu2023directed} introduces a novel approach for causal discovery in zero-inflated data, leveraging directed graphical models to enhance gene regulatory network analysis.
        Another example is \cite{zhao2024high}, which introduces a neighborhood selection method for learning the structure of Gaussian functional graphical models for high-dimensional functional data, applicable to EEG and fMRI data.
        Furthermore, there has been a growing interest in causal discovery for cyclic graphs \citep{richardson1996discovering, mooij2011causal, richardson2013discovery, sethuraman2023nodags}, as these models have implications for causal relationships that involve feedback loops and dynamic dependencies.
        It is noteworthy that cyclic structures pose additional challenges compared to acyclic graphs.

        Another avenue of research has concentrated on causal discovery methods that leverage interventional data as well as observational data \citep{kocaoglu2019characterization, brouillard2020differentiable, li2023causal}.
        Provided access to interventional data, one can reduce the size of the equivalence class, resulting in a finer specification of the causal model.
        Some works consider causal discovery in an active manner, where experiments are designed explicitly to learn causal graphs \citep{hyttinen2013experiment, hauser2014two, hu2014randomized, ghassami2017optimal, addanki2020efficient, mokhtarian2023unified}.
        This active approach involves strategically choosing interventions to gain the most informative data for learning the graphical structure.
        
        Some other notable lines of research include causal discovery for temporal data \citep{entner2010causal, assaad2022survey, chu2014causal, gong2023causal}, and causal representation learning \citep{scholkopf2021toward, wang2021desiderata, ahuja2023interventional}.

\section{RCD: A Python Package for Recursive Causal Discovery} \label{sec: RCD}
    We have implemented the algorithms presented along with other necessary and auxiliary utility functions in our Python package, called RCD.
    Our implementation is available on GitHub with the following link:
    \begin{center}
        \href{https://github.com/ban-epfl/rcd}{\url{github.com/ban-epfl/rcd}}
    \end{center}
    Additionally, you can find a detailed documentation of our package on the following website:
    \begin{center}
        \href{https://rcdpackage.com}{\url{rcdpackage.com}}
    \end{center}
    Some of the key aspects of our package are highlighted in the following.
    \begin{itemize}
        \item \textbf{Simple installation:} RCD is available on \texttt{PyPI} for installation.
        Use the command 
        \begin{center}
            \texttt{pip install rcd}
        \end{center}
        to add it to your environment.
        \item \textbf{Lightweight dependencies:} RCD uses only four packages - \texttt{NetworkX}, \texttt{NumPy}, \texttt{Pandas}, and \texttt{SciPy}, all commonly used in causal discovery.
        \item \textbf{Well-documented:} We have written thorough documentation for each class and function in the Google Python documentation style, available in the website.
        \item \textbf{Readable code:} We used consistent naming schemes for classes, functions, and variables, and added descriptive in-code comments for increased readability.
        \item \textbf{Efficient implementation:} Optimized for performance, we used optimal data structures and minimized loops and redundant CI tests, ensuring a lean and fast codebase.
    \end{itemize}

\subsection{Source Code Organization}
    The source code of the RCD package available on our GitHub repository is divided into four directories.
    \begin{itemize}
        \item \textbf{rcd}: It contains the implementation of the methods.
        \item \textbf{tests}: It contains unit tests in the framework of \texttt{pytest} that ensure the correctness of the implementation of each method.
        \item \textbf{examples}: It contains working demonstrations of each method.
        \item \textbf{docs}: It contains the configuration for \texttt{MkDocs}, which is responsible for generating our documentations site.
    \end{itemize}

\subsection{Method Implementation}
    Our proposed methods in the RCD package are implemented as Python classes with the following names:
    \begin{center}
        \texttt{Marvel},\quad \texttt{LMarvel}, \quad \texttt{RSLDiamondFree}, \quad \texttt{RSLBoundedClique}, \quad \texttt{ROLHillClimb}
    \end{center}
    RCD is designed with modularity at its core. Each class requires a CI testing function upon initialization and optionally accepts a Markov boundary matrix finding function. This design enables users to incorporate their own CI testing and Markov boundary-finding algorithms. If a Markov boundary-finding function is not given, our methods use a naive approach to find the initial Markov boundary matrix.
    
    Each instantiated class has a primary function that is intended to be used by the user, which is the \texttt{learn\textunderscore and\textunderscore get\textunderscore skeleton} function.
    This function receives a Pandas Dataframe as input, which contains data samples for each variable, with each column representing a variable.
    The function then returns the learned skeleton as an undirected NetworkX graph.

    Below is a small snippet showing the class corresponding to the RSL$_{D}$ method, named \texttt{RSLDiamondFree}.
    This class is instantiated and used to learn the skeleton.
    
    \begin{center}
        \begin{tabular}{l}
            \texttt{rsl\_d = RSLDiamondFree(custom\_ci\_test)} \\
            \texttt{learned\_skeleton = rsl\_d.learn\_and\_get\_skeleton(data\_df)}
        \end{tabular}
    \end{center}
    To see a fully functional demonstration, you can refer to the examples directory in the repository available at the following link:
    \begin{center}
        \href{https://github.com/ban-epfl/rcd/tree/main/examples}{github.com/ban-epfl/rcd/tree/main/examples}
    \end{center}
    Empirical evaluations of all of our methods on synthetically generated as well as real-world graphs can be referred to in their corresponding papers.




\section{Conclusion, Limitations, and Future Work}
    In this work, we developed a comprehensive framework for recursive causal discovery, derived from our previous publications (\refR{1}-\refR{4}), refined with additional details and enhancements.
    Our methodology revolves around identifying removable variables, learning their neighbors, discarding them, and then recursively learning the graph of the remaining variables.
    Through this iterative process, we have significantly reduced the number of performed CI tests, enhancing the computational efficiency and the accuracy of our methods.
    We further provided lower bounds on the complexity of constraint-based methods in the worst case and showed that our proposed methods almost match the lower bounds.
    Finally, we introduced \texttt{RCD}, a Python package that efficiently implements our methods.
    The main results discussed in this paper are summarized in Table \ref{table: summary of results}.
    
    \begin{table}[p]
        \centering
        \begin{adjustbox}{width=0.98\textwidth, totalheight=0.98\textheight, keepaspectratio}
            \begin{tabular}{N M{2.9cm}|M{8.5cm}|M{1.5cm}}
                \toprule
                & Result & Description & Source\\
                \hline
                & Proposition \ref{prp: removable}& Only removables can get removed & \refR{1}, \refR{2} \\
                & Theorem \ref{thm: graph-rep DAG} & Graphical characterization of removables in DAGs & \refR{1} \\
                & Theorem \ref{thm: graph-rep MAG} & Graphical characterization of removables in MAGs & \refR{2} \\
                & Proposition \ref{prp: removable exists} & Removables exist & \refR{1}, \refR{2} \\
                & Proposition \ref{prp: small Mb} & Removables have small Mb size & \refR{1}, \refR{2} \\
                & Proposition \ref{prp: invariant} & Removables are invariant in a MEC & \\
                & Proposition \ref{prp: r-order is learnable} & r-orders are invariant across MEC & \refR{4} \\
                & Proposition \ref{prp: c-o subset of r-o} & r-orders include c-orders & \refR{4} \\
                & Lemma \ref{lem: v-structures} & Finding v-structures & \refR{1}\\
                & Lemma \ref{lem: condition1} & Testing condition 1 of removability in DAGs & \refR{1}\\
                & Lemma \ref{lem: condition2} & Testing condition 2 of removability in DAGs & \refR{1}\\
                & Theorem \ref{thm: test removability} & Testing removability in MAGs & \refR{2}\\
                & Proposition \ref{prp: L-MARVEL duplicate} & Avoiding duplicate CI tests in L-MARVEL & \\
                & Theorem \ref{thm: FindRemovable clique} & Removability test in RSL$_{\omega}$ & \refR{3} \\
                & Proposition \ref{prp: FindNeighbors clique} & Finding neighbors in RSL$_{\omega}$ & \refR{3} \\
                & Theorem \ref{thm: FindRemovable RSL_D} & Removability test in RSL$_{D}$ & \refR{3} \\
                & Proposition \ref{prp: FindNeighbors RSL_D} & Finding neighbors in RSL$_{D}$ & \refR{3} \\
                & Theorem \ref{thm: opt prob} & Consistency result of ROL's objective & \refR{4} \\
                & Proposition \ref{prp: verifiable} & RSL$_{\omega}$ is verifiable & \refR{3} \\
                & Theorem \ref{thm: lower Bound DAG} & Lower bound for DAGs & \refR{1} \\
                & Theorem \ref{thm: lower Bound MAG} & Lower bound for MAGs & \refR{2} \\
                & Proposition \ref{prp: complexity of MARVEL} & Completeness and complexity of MARVEL & \refR{1} \\
                & Proposition \ref{prp: complexity of L-MARVEL} & Completeness and complexity of L-MARVEL & \refR{2} \\
                & Proposition \ref{prp: complexity of RSL_w} & Completeness and complexity of RSL$_{\omega}$ & \refR{3} \\
                & Proposition \ref{prp: complexity of RSL_D} & Completeness and complexity of RSL$_{D}$ & \refR{3} \\
                & Proposition \ref{prp: complexity of ROL_HC} & Complexity of ROL$_{HC}$ &  \\
                & Proposition \ref{prp: complexity of ROL_VI} & Completeness and complexity of ROL$_{VI}$ & \refR{4} \\            
            \bottomrule
            \end{tabular}
        \end{adjustbox}
        \caption{Table of results.}
        \label{table: summary of results}
    \end{table}

    While our framework offers significant advancements in recursive causal discovery, it shares a common limitation with all constraint-based methods: it relies on CI tests, which may not consistently provide reliable results, especially in non-parametric models where such tests can be less robust.
    Additionally, our approaches leverage Markov boundaries in their recursive framework, allowing the application of any existing algorithm for computing Markov boundaries.
    Although we proposed methods that iteratively update the Markov boundaries at a low cost, the initial step may pose a computational bottleneck in large-scale applications.

    In the following, we discuss potential future work.
    \begin{itemize}
        \item
            As mentioned earlier, although our current approaches can learn graphs up to the order of $10^3$ variables with conventional computational power, the initial computation of Markov boundaries poses a computational challenge for larger graphs.
            An influential direction for future work is the development of recursive causal discovery methods that do not depend on Markov boundary computations.
        \item
            In exploring real-world applications of our package, one domain of particular interest is network biology, specifically for learning Gene Regulatory Networks (GRN).
            However, given the large number of genes in the human genome (approximately 20,000), applying our current methods directly might be challenging due to the dimensionality issues.
            While our methods can be used to analyze GRNs of other organisms with fewer genes or adapted for local analyses within the human genome, future modifications aimed at tailoring our recursive causal discovery methods for high-dimensional biology data, such as GRNs, could unlock new opportunities for applying our package to complex biological systems.
        \item
            Another direction for future work involves the parallelization of our proposed recursive causal discovery methods.
            By leveraging parallel computing techniques, we can significantly reduce the computational time, making our algorithms more efficient and scalable.
        \item 
            Another research question is to analyze the sample complexity of our proposed recursive causal discovery methods.
            While our methods have empirically shown to require fewer samples compared to other approaches in \refR{1}-\refR{4}, a theoretical analysis of sample complexity remains an open question.
    \end{itemize}

\bibliography{bibliography}

\clearpage
\end{document}